\documentclass[sn-mathphys]{sn-jnl}

\usepackage{graphicx}%
\usepackage{subcaption}
\usepackage{multirow}%
\usepackage{amsmath,amssymb,amsfonts}%
\usepackage{amsthm}%
\usepackage{mathrsfs}%
\usepackage[title]{appendix}%
\usepackage{xcolor}%
\usepackage{textcomp}%
\usepackage{manyfoot}%
\usepackage{booktabs}%
\usepackage{algorithm}%
\usepackage{algorithmicx}%
\usepackage{algpseudocode}%
\usepackage{listings}%
\usepackage{todonotes}%
\usepackage{color,soul}
\usepackage{svg}

\begin{document}

% npj-style: keep <= 15 words, no colon, no puns.
\title[Expressive swarm robotics]{Reality and Perceptual Gaps in Expressive Swarm Robotics}
%\title[Expressive swarm robotics]{Bridging reality and perceptual gaps in expressive swarm robotics}

%%%%=========%%
\author[1]{\fnm{Ali} \sur{Imran}}
\author[1]{\fnm{Abdalwhab Bakheet Mohammed} \sur{Abdalwhab}}
\author[1]{\fnm{Rafael} \sur{Gomes Braga}}
\author[3]{\fnm{Hélène} \sur{Duval}}
\author[1]{\fnm{Audrey} \sur{Rochette}}
\author[2]{\fnm{Vivek Shankar} \sur{Varadharajan}}
\author[2]{\fnm{Yann} \sur{Bouteiller}}
\author[1]{\fnm{Jean} \sur{Mazerolle}}
\author[2]{\fnm{Giovanni} \sur{Beltrame}}
\author*[1]{\fnm{David} \sur{St-Onge}}
% \email{adresse@etsmtl.ca}

%% Affiliations
\affil[1]{
    \orgdiv{INIT Robots Laboratory, Department of Mechanical Engineering},
    \orgname{École de technologie supérieure},
    \orgaddress{
        \city{Montréal},
        \state{Québec},
        \country{Canada}
    }
}

\affil[2]{
    \orgdiv{MIST Lab, Department of Computer and Software Engineering},
    \orgname{Polytechnique Montréal},
    \orgaddress{
        \city{Montréal},
        \state{Québec},
        \country{Canada}
    }
}

\affil[3]{
    \orgdiv{Department of Dance},
    \orgname{Université du Québec à Montréal},
    \orgaddress{
        \city{Montréal},
        \state{Québec},
        \country{Canada}
    }
}
%%==================================%%

\abstract{%

In swarm robotics, the reality gap arises when discrepancies in sensing, timing and actuation alter collective behaviour during transfer from simulation to physical deployment. In human-facing settings, a second challenge arises: the perceptual gap between realised collective organisation and human interpretation. We investigate both gaps through a research-creation project deploying shared decentralised behavioural code across tabletop, aerial and human-scale robotic swarms in public performances. Fireworks and Pursuit preserved their defining collective organisation across embodiments, yet their motion differed in geometry, directional structure and timing, with Fireworks periods varying more than threefold. Audience questionnaires (54 respondents) and qualitative responses revealed differences across embodiments in perceived meaning, sympathy, hostility and comfort. Meaning attribution did not necessarily coincide with positive affect. These findings distinguish the successful physical realisation of collective behaviour from its human interpretation and highlight the need to consider expressive motion alongside technical deployment performance in human-facing swarm robotics.}


\maketitle

%%%%%%%%%%%%%%%%%%%%%%%%%%%%%%%%%%%%%%%%%%%%%%%%%%%%%%%%%%%%%%%%%%%%%%
\section{Introduction}
%%%%%%%%%%%%%%%%%%%%%%%%%%%%%%%%%%%%%%%%%%%%%%%%%%%%%%%%%%%%%%%%%%%%%%

Over the past decades, swarm robotics has moved from predominantly simulation-driven studies toward increasingly capable physical multi-robot systems. Software frameworks such as ROS2swarm~\cite{kaiser2022ros2swarm}, CrazyChoir~\cite{pichierri2023crazychoir}, Crazyswarm2~\cite{crazyswarm2}, and Aerostack2~\cite{fernandez2023aerostack2} have made collective behaviours more modular across specific hardware ecosystems, while large field demonstrations have shown that multi-robot coordination can persist in unstructured environments~\cite{rouvcek2020darpa}. For swarms, however, transferring a controller from simulation to hardware is more than a conventional sim-to-real problem repeated across many robots. Collective behaviour emerges from repeated local interactions, so small discrepancies in sensing, localisation, communication timing, actuation, or update frequency can propagate through the network and alter group-level dynamics, even when the same decentralised program executes correctly on every agent. The transfer target is therefore not only an agent-level controller but also the interaction process from which the collective regime emerges. A behaviour may therefore survive deployment at the code level while changing in period, spatial organisation, convergence, or interaction structure at the collective level. Existing work has addressed these effects through simulator-in-the-loop refinement~\cite{vega2023simulate, ju2022review}, Real2Sim2Real transfer of emergent behaviours~\cite{mattson2025discovery}, mixed-reality validation over physical communication networks~\cite{sende2024bridging}, and quantitative robustness and deployment metrics~\cite{bozhinoski2023enhancing, robotswarmrealworld2022, milner2023swarm}. Across much of this work, transfer is assessed from the behaviour of the robotic system itself: whether the physical swarm realises the collective organisation specified by its controller.

\begin{figure}[ht]
  \centering
  \includesvg[width=\columnwidth]{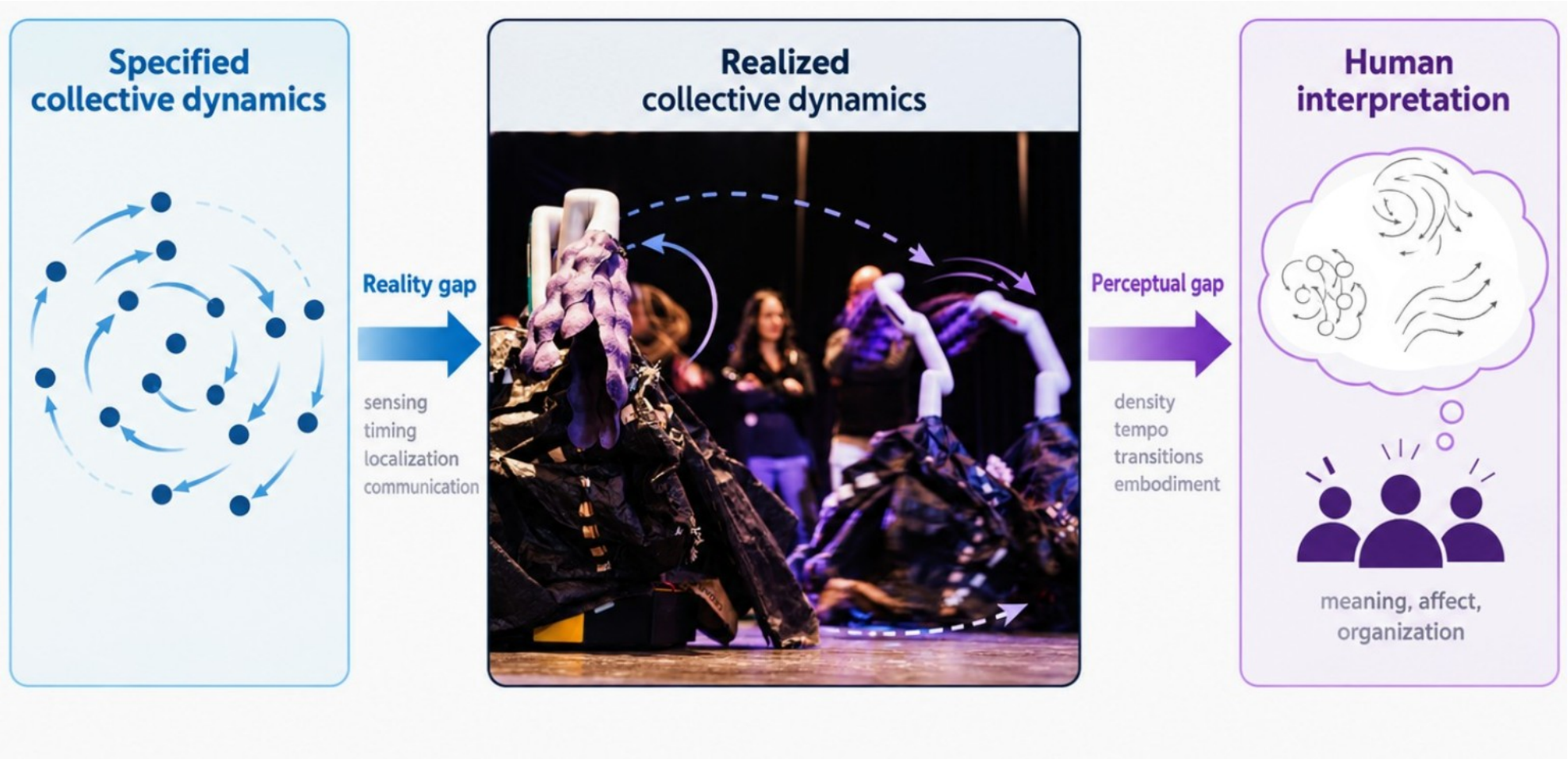}
  \caption{From specified collective organisation to human interpretation: the sequential reality and perceptual gaps examined in this work. The physically realised collective forms the stimulus available to human perception.}
  \label{fig:optivsuwb}
\end{figure}

%ARGoS~\cite{pinciroli2012argos} or Webots~\cite{Webots2022} degrade in the field under sensing noise, communication loss, occlusion, and interference that simulation rarely captures~\cite{vega2023simulate, ju2022review}. These issues have been addressed with simulator-in-the-loop refinement~\cite{vega2023simulate}, Real2Sim2Real transfer of emergent behaviors~\cite{mattson2025discovery}, mixed-reality validation of drone swarms over live networks~\cite{sende2024bridging}, and benchmarking, and quantitative robustness metrics~\cite{bozhinoski2023enhancing, robotswarmrealworld2022, milner2023swarm}. Underlying all of this work is a shared success criterion: a deployment succeeds when the swarm executes, on hardware, the collective behavior its scripts specify.

Two years of deploying heterogeneous swarms in public performances exposed a complementary problem: a collective behaviour can be physically realised as specified and still remain difficult for a human observer to interpret. We refer to the discrepancy between the collective organisation physically realised by the swarm and the organisation inferred by an observer as the \emph{perceptual gap}. This challenge becomes increasingly relevant as robot collectives move into spaces shared with non-expert humans, where motion must support anticipation, prediction, and appropriate trust~\cite{chen2014human,dorigo2021swarm,hepworth2020human}.

% \hl{We argue that public-facing swarms confront a second gap, which we call the \textit{perceptual gap}: the discrepancy between the collective organisation a designer intends and the collective aim an observer perceives. The gap has two legs, one from intended organisation to the collective that physically appears, and one from that collective to what an observer reads.}

As illustrated in Fig.~\ref{fig:optivsuwb}, the perceptual gap follows the reality gap rather than replacing it. A designer specifies collective dynamics, which robots physically realise through interactions shaped by sensing, timing, and embodiment. This realised motion then becomes the stimulus from which human observers interpret the collective. The two gaps are therefore distinct but sequentially coupled: discrepancies introduced during physical realisation shape what becomes available to perception, while preserving the specified collective organisation does not guarantee how it will be interpreted. This distinction motivates characterising the collective motion realised across robotic embodiments before examining audience interpretations.

The perceptual gap in a swarm is not simply the legibility problem of a single robot multiplied by the number of agents. Classical legibility concerns inference of an intended goal from observed motion~\cite{dragan2013legibility, dragan2015effects}. Recent work on multi-robot systems extends this notion to coordination objectives—what the group is doing relative to itself—as well as spatial goals~\cite{capelli2024}. For expressive swarms, however, the perceptual question is broader: collective behaviours need not be goal-directed, and observers may construct meaning from their organisation without inferring a predefined objective. In a decentralised swarm, the organisation of interest often exists primarily in the relations among agents. Individual robots execute local rules and exchange information that remains invisible to an external observer; the collective can therefore be read through changes in density, alignment, synchrony, spatial distribution, transitions between behavioural regimes, and the evolving relations among agents. Embodiment and context further shape this reading through scale, morphology, sound, appearance, and spatial framing. Existing work already indicates that collective-behaviour cues can be more informative than trajectory-level cues for swarm legibility in goal-oriented tasks~\cite{kim2021generating}, and that spatial distribution and density influence perceived agency in embodied swarms~\cite{ichihashi2024swarm}. Human--swarm transparency research has mainly addressed operators interacting through dedicated interfaces~\cite{patel2022multi,kolling2015human}, whereas public-facing observers must make sense of the collective from its behaviour alone. In parallel, expressive-motion approaches based on Laban analysis~\cite{Burton2016}, animation pipelines~\cite{Moioli2022}, or generative models~\cite{mahadevan2024generative} have largely focused on individual robots. How expressive intent can be observed, composed, and iteratively refined at the level of a decentralised collective therefore remains comparatively unexplored, particularly outside artistic research-creation settings~\cite{naoko2022,cuan2020}.

To investigate these two gaps together, we draw on DESSAIM, an interdisciplinary research-creation project bringing together artists and roboticists to develop and stage expressive collective behaviours. The project brought together three robotic swarms spanning markedly different physical scales: tabletop Sushis, aerial CrazyCognies, and human-scale mobile manipulators, the Doodies (Sec.~\ref{subsec:Platforms}). Across these systems, the same Buzz behavioural code~\cite{pinciroli2016buzz} was deployed through PyBuzz, a software stack that confines platform-specific sensing and actuation to adapter layers (Sec.~\ref{subsec:pybuzz}). This common representation enabled comparison of how the same collective behaviours were physically realised across embodiments, while exposing a limitation of platform abstraction: parameters expressed in controller steps may retain their numerical values without preserving their physical timescales. 

On the perceptual side, DESSAIM adopted an iterative research-creation process combining behaviour composition, transitions between swarm states, Movement Observation--Analysis (MOA)~\cite{harbonnier2021}, embodiment, scenography, sound, and repeated public presentations and surveys. These elements were co-designed rather than manipulated as independent experimental factors, forming a coupled expressive design space through which artists and roboticists could observe how collective motion was interpreted and guide subsequent revisions. Audience questionnaires and open-ended responses collected across successive performances provide complementary evidence of how meaning, affect, agency, and collective organisation were attributed differently to the three embodiments. In particular, the results indicate that meaning attribution and affect need not vary together: collective motion may be interpreted as meaningful while simultaneously eliciting feelings of discomfort or threat.

The purpose of this work is therefore narrower than a general report on field
deployment, but broader than the validation of a software framework. PyBuzz,
localisation, communication, and cross-platform behavioural analysis are used
here as engineering instruments for exposing the simulation-to-deployment gap
rather than as the subject of a standalone software contribution. Conversely,
the artistic practices are not treated as an art-theoretical endpoint, but as
methods for interrogating how collective robotic motion is composed, observed,
and interpreted. Public performance provides an ecological testbed in which
these concerns meet: the swarm must remain technically coherent under changing
physical conditions while also remaining available to human perception and
interpretation.

Taken together, the study combines a quantitative cross-platform
characterisation of shared swarm behaviours, localisation results under
performance constraints, and audience-perception data with a two-axis
methodology for working on the reality and perceptual gaps. The central argument
is not that either gap has been closed, but that human-facing swarm robotics
must make both observable and actionable. Robust deployment infrastructure is
required to preserve collective organisation across embodiments; an expressive
toolkit is required to understand and shape how that organisation is perceived
and interpreted once it reaches the physical world.

%\todo[inline]{do we need a para for paper structure explanantion? and contributions of the paper?} - contributions are mentionned. structure doesn't seem to be in NPJ papers...
%%%%%%%%%%%%%%%%%%%%%%%%%%%%%%%%%%%%%%%%%%%%%%%%%%%%%%%%%%%%%%%%%%%%%%

\section{Results}\label{sec:results}
%%%%%%%%%%%%%%%%%%%%%%%%%%%%%%%%%%%%%%%%%%%%%%%%%%%%%%%%%%%%%%%%%%%%%%

\begin{figure*}[htp]
    \centering
    \subcaptionbox{Tabletop swarm: Sushis\label{fig:sushis}}{%
        \includegraphics[trim={0 30cm 0 0},clip,height=1.7in]{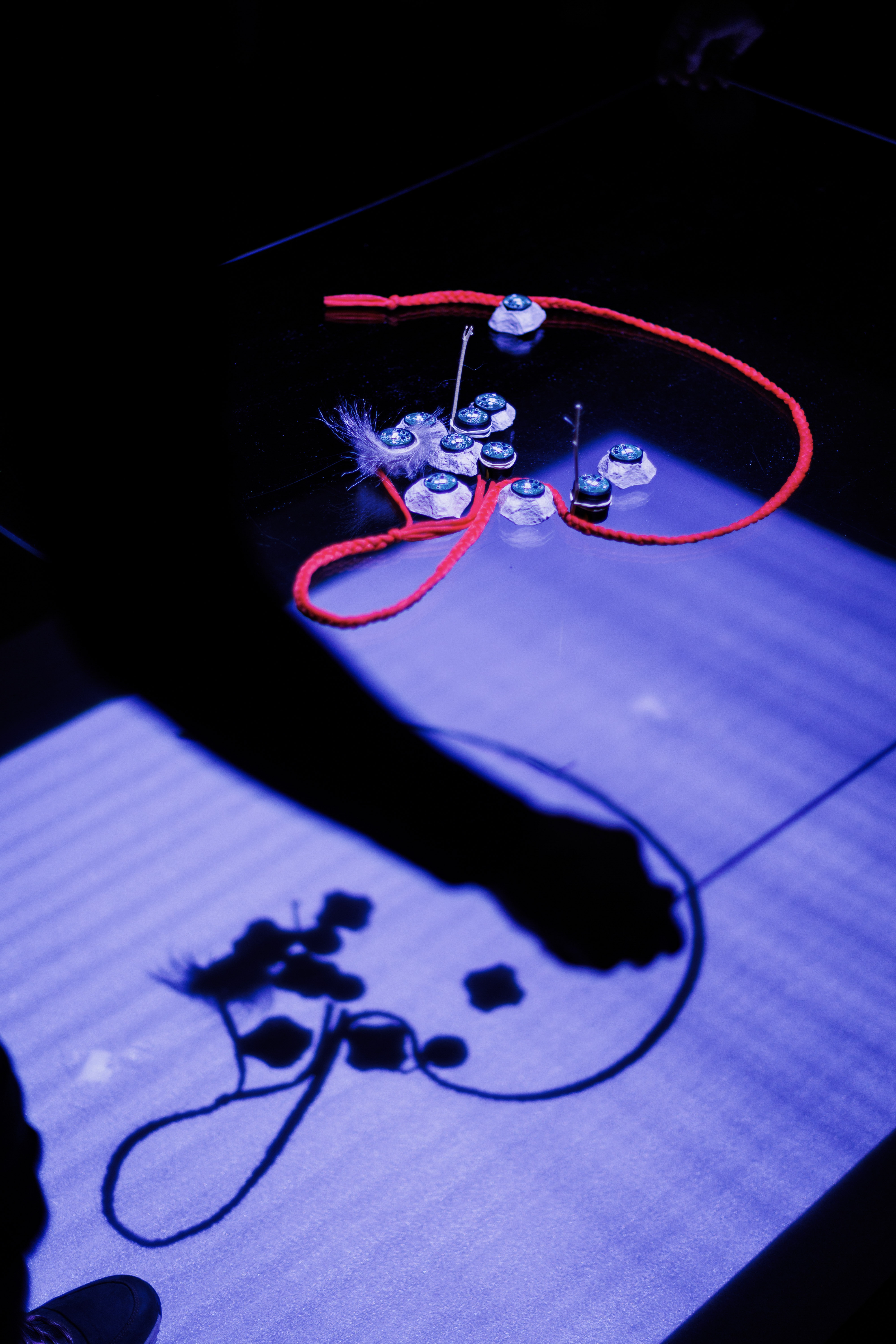}%
    }\quad
    \subcaptionbox{Flying swarm: CrazyCognies\label{fig:cc}}{%
        \includegraphics[height=1.7in]{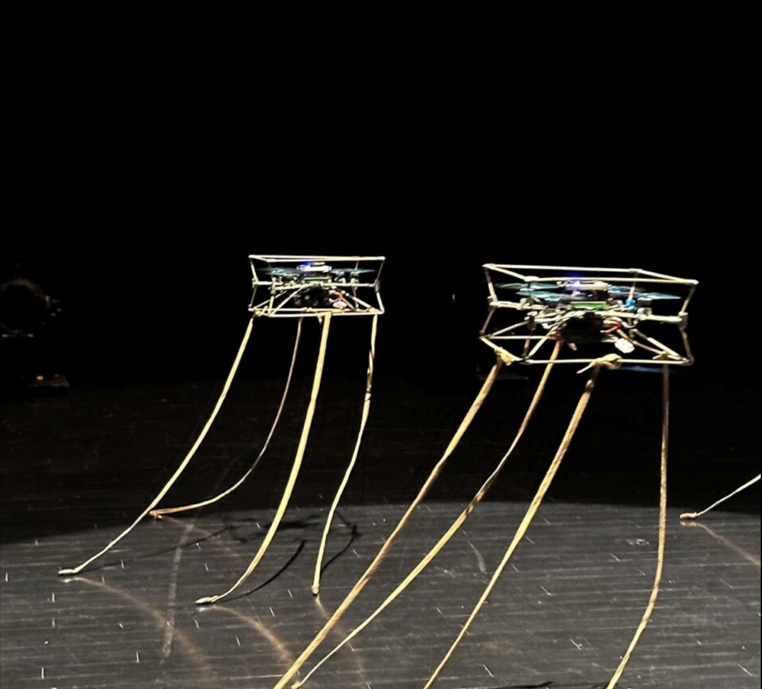}%
    }\quad
    \subcaptionbox{Human-scale swarm: Doodies\label{fig:Doodies}}{%
        \includegraphics[height=1.7in]{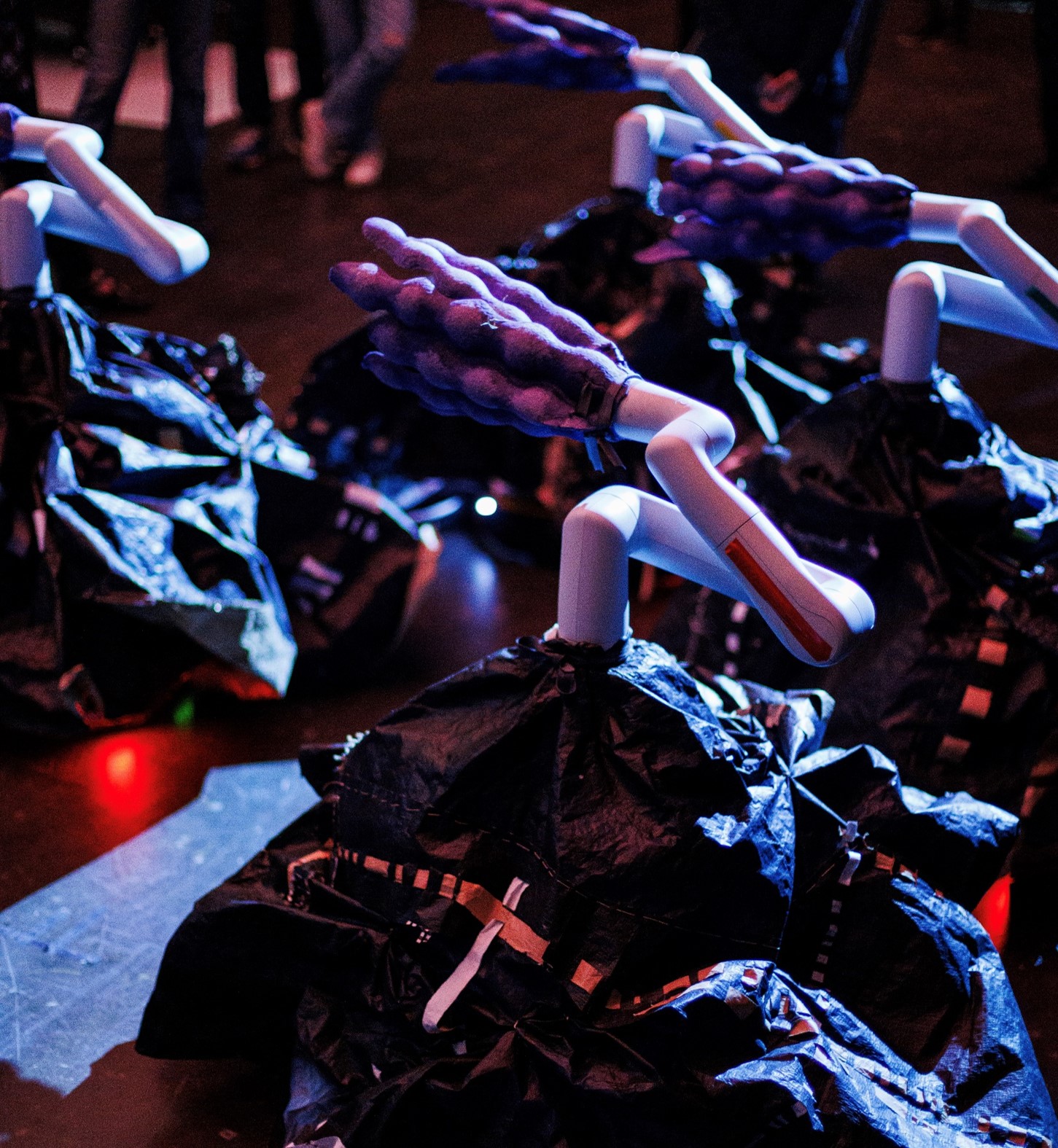}%
    }
    \caption{The three DESSAIM robotic swarms seen in their performing settings.\label{fig:platforms}}
\end{figure*}

The DESSAIM performances brought together three robotic collectives, shown in
Fig.~\ref{fig:platforms}, spanning markedly different scales, dynamics, and
swarm sizes: tabletop Sushis, aerial CrazyCognies, and human-scale mobile
manipulators, the Doodies. These heterogeneous embodiments provided a common
setting in which to examine both steps of that chain: what a collective
behaviour designed in simulation or software becomes once a particular body
realises it, and how the realised collective is then read by human observers.
Our results therefore focus on three bodies of evidence: the physical
realisation of common emergent behaviours across embodiments, localisation
performance during expressive motion, and quantitative and qualitative
audience-perception outcomes. The software, localisation, composition, staging,
and audience-study procedures used to work on these two gaps are described in
Methods (Sec.~\ref{sec:methods}).

\subsection{Emergent behaviours across platforms}\label{subsec:EmergentBehaviours}

\begin{figure*}[!htb]
\centering
\includegraphics[width=0.95\textwidth]{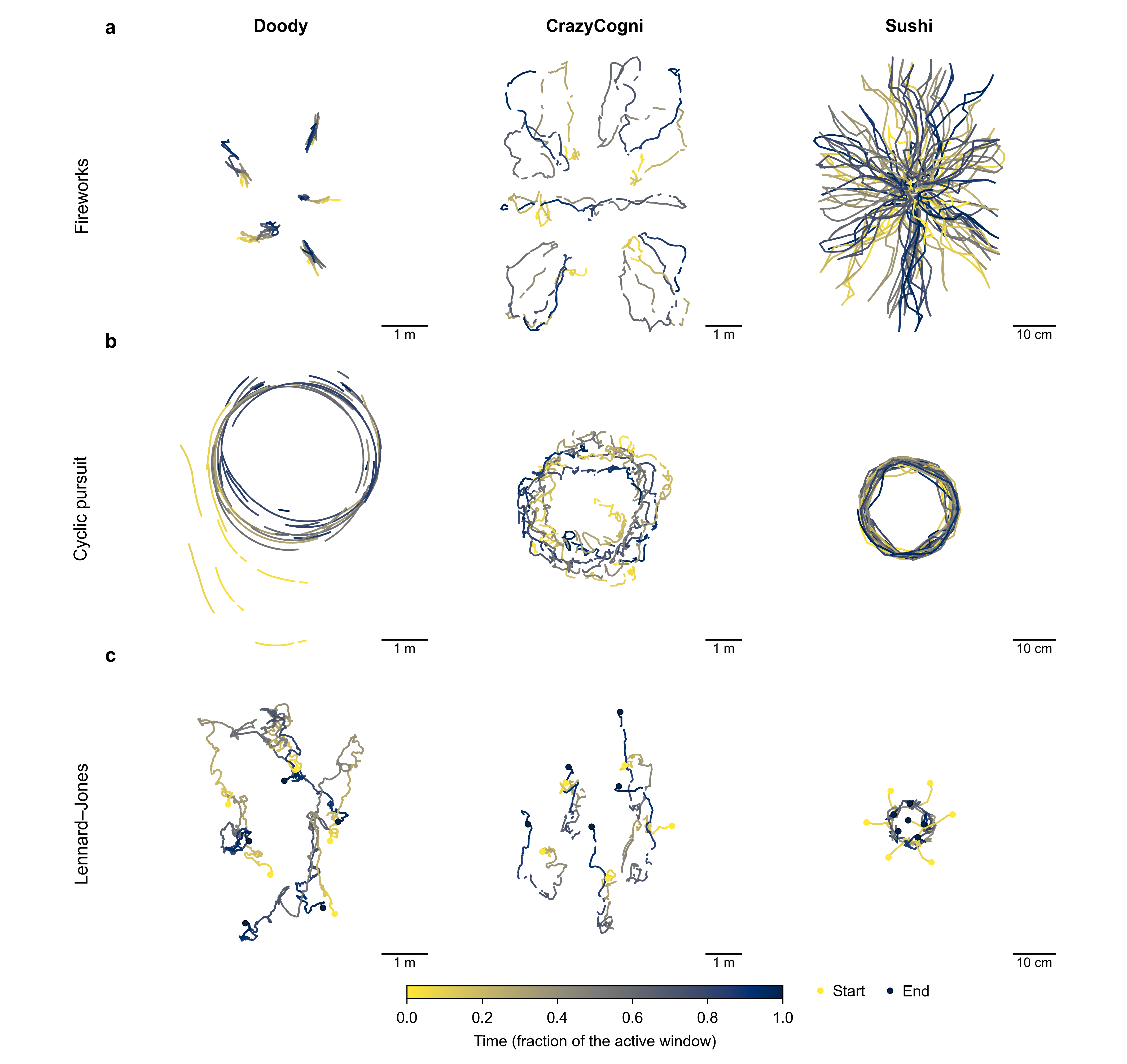}
% \caption{\textbf{One script, three pictures.} 
\caption{Example trajectories of the three behaviours on each platform: (a) Fireworks, (b) cyclic Pursuit, and (c) Lennard--Jones. Each behaviour runs from the same script on the Doodies, CrazyCognies and Sushis. Lines show individual robots, coloured by time.}
\label{fig:s21traj}
\end{figure*}

Artists and roboticists jointly explored canonical swarm behaviours, including aggregation, diffusion, flocking~\cite{schranz2020swarm}, Lennard--Jones interactions~\cite{rosbuzz2020}, and pursuit~\cite{pursuit2018}. These primitives formed the basis of a twelve-behaviour expressive library developed through the iterative creation process (Sec.~\ref{sec:library}). The complete library was performed with the Sushis, whereas Lennard--Jones, Fireworks, and Pursuit formed the common subset deployed on all three robotic embodiments and analysed here.

Running the same behaviours on three bodies allowed us to ask whether each behaviour’s defining collective structure survived physical embodiment, and how its spatiotemporal characteristics changed across platforms. We quantified externally observable trajectory geometry, spatial scale, directional structure and characteristic timing from the robots' positions and velocities relative to the swarm centroid. Controller state, timing counters, operator inputs, and radio traffic were excluded because they are not directly observable from the realised collective motion.

For each of the three behaviours shared across platforms, we recorded three takes per embodiment. External trajectories were obtained by motion capture for the Doodies and CrazyCognies and by overhead video for the Sushis in a separate bench session. For the CrazyCognies, we additionally retained the pose estimated onboard and broadcast between robots. Unless otherwise stated, the cross-platform results are reported from the external trajectories, which represent the motion physically realised by the swarm. The onboard CrazyCogni estimates are used only to test whether conclusions drawn from the external motion persist when the same descriptors are computed from the robots’ own state
estimates.

The shared behaviours produced recognisable group-level dynamics while taking markedly different physical forms (Fig.~\ref{fig:s21traj}). Fireworks drew radial strokes on the Doodies and the Sushis, with each robot moving away from the group and back along roughly the same line. On the CrazyCognies, it drew loops, and the outward and return paths rarely coincided. Pursuit drew a circle on every body, but the 0.11~m orbit that the Sushis completed every 7.9~s became a 1.4--1.6~m orbit taking 58~s on the Doodies and about 100~s on the CrazyCognies (Table~\ref{tab:realised}). Lennard--Jones gave three different collectives: a slowly drifting lattice on the Doodies, a restless cloud on the CrazyCognies that moved as fast as their pursuit, and, on the Sushis, a near-static formation with approximately uniform spacing, interrupted only by occasional corrective movements.

\begin{figure*}[!htb]
\centering
\includegraphics[width=0.95\textwidth]{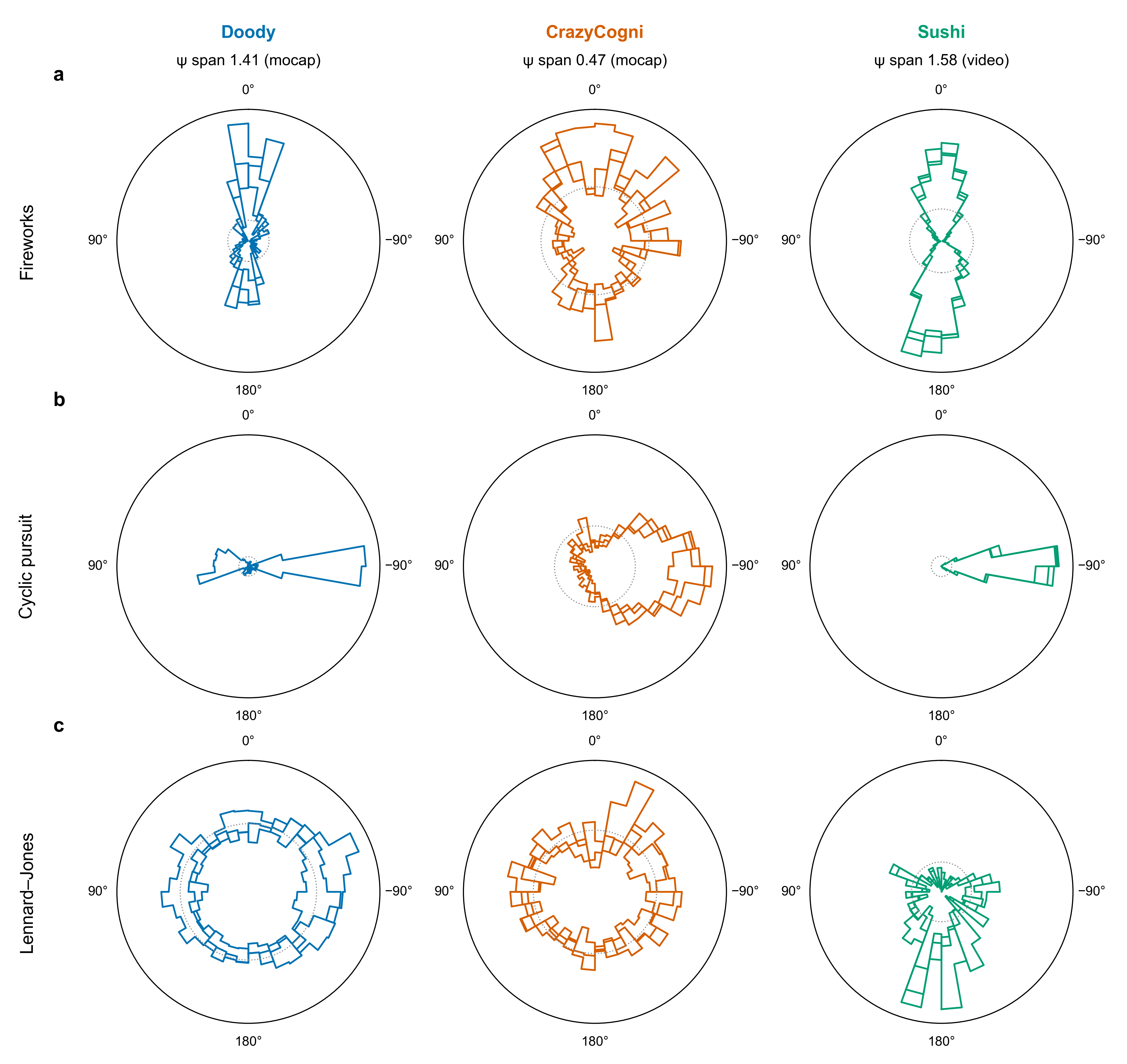}
\caption{Distribution of each robot's direction of motion relative to the outward radial direction from the swarm centroid, $\alpha$ (0\textdegree{} outward, 180\textdegree{} inward, $\pm$90\textdegree{} tangential), for Fireworks (\textbf{a}), cyclic Pursuit (\textbf{b}), and Lennard--Jones (\textbf{c}) on the Doodies (motion capture), CrazyCognies (motion capture), and Sushis (overhead video). Radial motion produces a vertical bow-tie distribution, sustained circulation produces a lateral lobe, and motion without a preferred radial or tangential orientation produces an approximately circular distribution. Each outline represents one take; the dotted circle denotes a uniform angular distribution, and the polar-density scale is set independently for each panel. Column headers report the $\psi$ span between the most radial Fireworks take and the most tangential Pursuit take.}
\label{fig:s21bowtie}
\end{figure*}

To compare the directional structure of the behaviours across embodiments, we examined each robot's heading relative to its outward direction from the swarm centroid (Fig.~\ref{fig:s21bowtie}). Fireworks produced the expected radial bow-tie pattern, Pursuit a tangential lobe, and Lennard–Jones produced an approximately circular distribution of movement headings (six units produce an hexagon). The radial index $\psi$ (Eq.~\ref{eq:psi}; Section~\ref{subsec:analysis}) captured the same ordering across platforms and measurement layers, with Fireworks the most radial and Pursuit the most tangential (Table~\ref{tab:realised}). Fireworks also retained its cyclic structure: for every take containing at least three complete cycles, the estimated period varied by only 0.2--2.6\% across detrending windows (Supplementary Note~1). Together, these results satisfy the transfer criterion defined in Section~\ref{subsec:analysis} for Fireworks and Pursuit on all three platforms.

Directional separation was less well preserved across embodiments. The Fireworks and Pursuit signatures remained sharply separated on the Doodies and Sushis but were compressed on the CrazyCognies (Fig.~\ref{fig:s21bowtie}, Table~\ref{tab:realised}). Between the most radial Fireworks take and the most tangential Pursuit take, $\psi$ spanned 1.41 on the Doodies and 1.58 on the Sushis, but only 0.47 on the CrazyCognies. The same compression was evident within Pursuit: $\psi$ was $-0.19\pm0.05$ on the CrazyCognies, compared with $-0.61\pm0.22$ on the Doodies and $-0.896\pm0.012$ on the Sushis, while the settled milling order reached only 0.50--0.51 compared with 0.98--1.00 on the other two embodiments (Table~\ref{tab:realised}). Because both the smallest and largest embodiments retained substantially wider separation, this pattern is not consistent with a simple effect of physical scale. The aerial swarm therefore realised the three behaviours with markedly weaker directional differentiation.

\begin{figure*}[!t]
\centering
\includegraphics[width=0.95\textwidth]{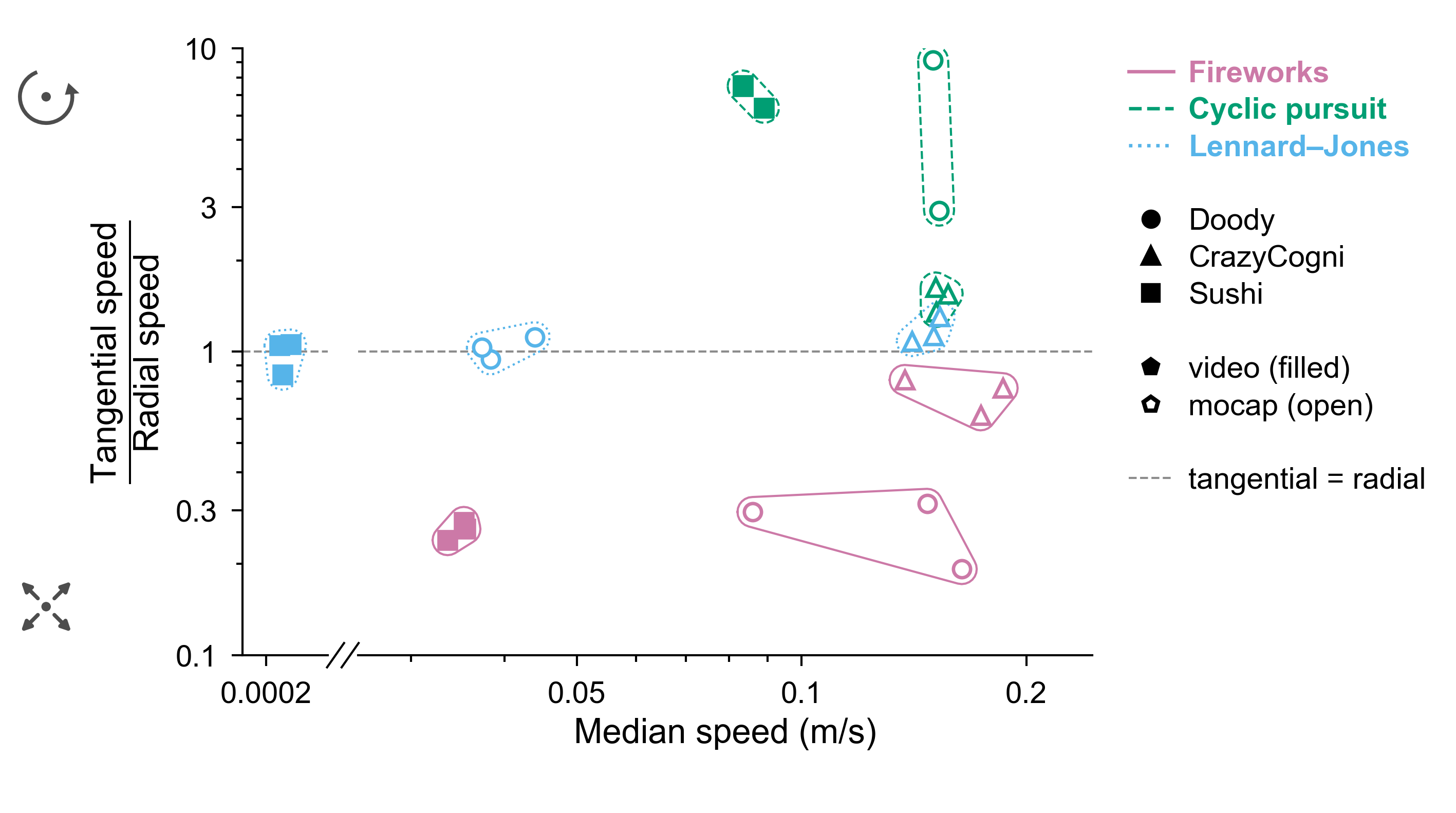}
\caption{Median robot speed against the ratio of median tangential to median radial speed relative to the swarm centroid, on logarithmic axes. Above the dashed line, motion is predominantly tangential; below it, predominantly radial. Outlines group takes from the same platform--behaviour combination. Left of the axis break are the three Sushi Lennard--Jones takes; their very low speeds mean that the ratio is dominated by near-stationary periods rather than the brief corrective motions that maintained spacing.}
\label{fig:s21velocity}
\end{figure*}

Figure~\ref{fig:s21velocity} compares the overall speed of each take with its
balance between tangential and radial motion. Fireworks remained radially
dominated on all three embodiments, Pursuit tangentially dominated, and
Lennard--Jones remained close to equal radial and tangential components. The
directional separation was again weaker on the CrazyCognies. During Fireworks,
radial speed exceeded tangential speed by factors of 3.2--5.2 on the Doodies and
3.6--4.2 on the Sushis, but only 1.2--1.6 on the CrazyCognies. Pursuit showed
the same compression: tangential speed exceeded radial speed by 2.9--9.1 on the
Doodies and 6.4--7.5 on the Sushis, compared with only 1.4--1.6 on the
CrazyCognies. Overall speed followed a different pattern. The CrazyCognies moved
at 148--166~mm\,s$^{-1}$ across all three behaviours, whereas Lennard--Jones
slowed the Doodies to 40~mm\,s$^{-1}$ and brought the Sushis close to rest
(Table~\ref{tab:realised}). This slowing is consistent with convergence toward
the target inter-agent spacing, where the Lennard--Jones interaction approaches
equilibrium; by contrast, the CrazyCognies retained speeds comparable to Pursuit
and remained a continuously moving cloud.

\begin{table}[!ht]
\caption{$R_g$, radius
of gyration; $\psi$, radial index; $P$, polarisation; $M$, milling order; modulation
depth, peak-to-trough swing of the phase-folded $R_g$ over its median. Measurements were obtained from motion capture for the Doodies and CrazyCognies and overhead video for the Sushis. Dashes mark values that are undefined: during Lennard--Jones on the Sushis, no robot moved fast enough to have a heading at 78--83\% of time steps, so $P$ and $M$ could not be computed.}
\label{tab:realised}
\small
\begin{tabular*}{\linewidth}{@{\extracolsep{\fill}}lccc@{}}
\toprule
 & \textbf{Doody} & \textbf{CrazyCogni} & \textbf{Sushi} \\
\midrule
\multicolumn{4}{@{}l}{\textit{Fireworks}} \\
Median $R_g$ (m)             & $1.45\pm0.24$    & $1.53\pm0.05$  & $0.166\pm0.004$ \\
Median speed (mm\,s$^{-1}$)  & $130\pm40$       & $166\pm25$         & $34.8\pm1.0$ \\
$\psi$                       & $+0.57\pm0.07$   & $+0.16\pm0.07$ & $+0.653\pm0.015$ \\
$P$                          & $0.33\pm0.08$    & $0.29\pm0.07$      & $0.157\pm0.015$ \\
$M$                          & $0.175\pm0.029$  & $0.17\pm0.04$      & $0.055\pm0.008$ \\
\addlinespace
\multicolumn{4}{@{}l}{\textit{Cyclic pursuit}} \\
Median $R_g$ (m)             & $1.4\pm0.4$      & $1.601\pm0.008$    & $0.1061\pm0.0029$ \\
Median speed (mm\,s$^{-1}$)  & $151.4\pm2.0$    & $153.2\pm3.3$      & $85.3\pm3.2$ \\
$\psi$                       & $-0.61\pm0.22$   & $-0.19\pm0.05$     & $-0.896\pm0.012$ \\
$P$                          & $0.4\pm0.4$      & $0.293\pm0.021$    & $0.11\pm0.06$ \\
$M$                          & $0.7\pm0.4$  & $0.45\pm0.06$  & $0.9768\pm0.0034$ \\
\addlinespace
\multicolumn{4}{@{}l}{\textit{Lennard--Jones}} \\
Median $R_g$ (m)             & $1.469\pm0.015$  & $1.48\pm0.13$      & $0.04137\pm0.00034$ \\
Median speed (mm\,s$^{-1}$)  & $40\pm4$         & $148\pm7$          & $0.30\pm0.04$ \\
$\psi$                       & $-0.023\pm0.035$ & $-0.06\pm0.05$     & $+0.05\pm0.07$ \\
$P$                          & $0.376\pm0.017$  & $0.47\pm0.08$      & -- \\
$M$                          & $0.190\pm0.021$  & $0.158\pm0.026$    & -- \\
\midrule
\multicolumn{4}{@{}l}{\textit{Behaviour-specific}} \\
Fireworks cycle period (s)   & $7.58\pm0.09$    & $26.9$         & $12.036\pm0.023$ \\
Fireworks modulation depth   & $0.27\pm0.06$    & $0.79$         & $1.335\pm0.027$ \\
Pursuit orbital period (s)   & $58\pm18$        & $101\pm10$     & $7.87\pm0.07$ \\
\bottomrule
\end{tabular*}
\end{table}

Characteristic timing also differed substantially across embodiments. The
Fireworks cycle lasted 7.6~s on the Doodies, 12.0~s on the Sushis, and 26.9~s on
the CrazyCognies, corresponding to a 3.5-fold spread for the same control law
(Table~\ref{tab:realised}). The Fireworks transition is triggered by a counter
that advances once per control step and uses no platform-specific timing
constant (Section~\ref{sec:library}), while PyBuzz does not impose a common
physical time base across platforms. On the Doodies and CrazyCognies, one cycle
required essentially the same number of control steps (157--161 and 158,
respectively), but the controller ran at 21~Hz on the Doodies and 5.9~Hz on the
CrazyCognies. The resulting loop-rate ratio (3.57) closely matched the measured
period ratio (3.55).

\begin{figure*}[!t]
\centering
\includegraphics[width=0.85\textwidth]{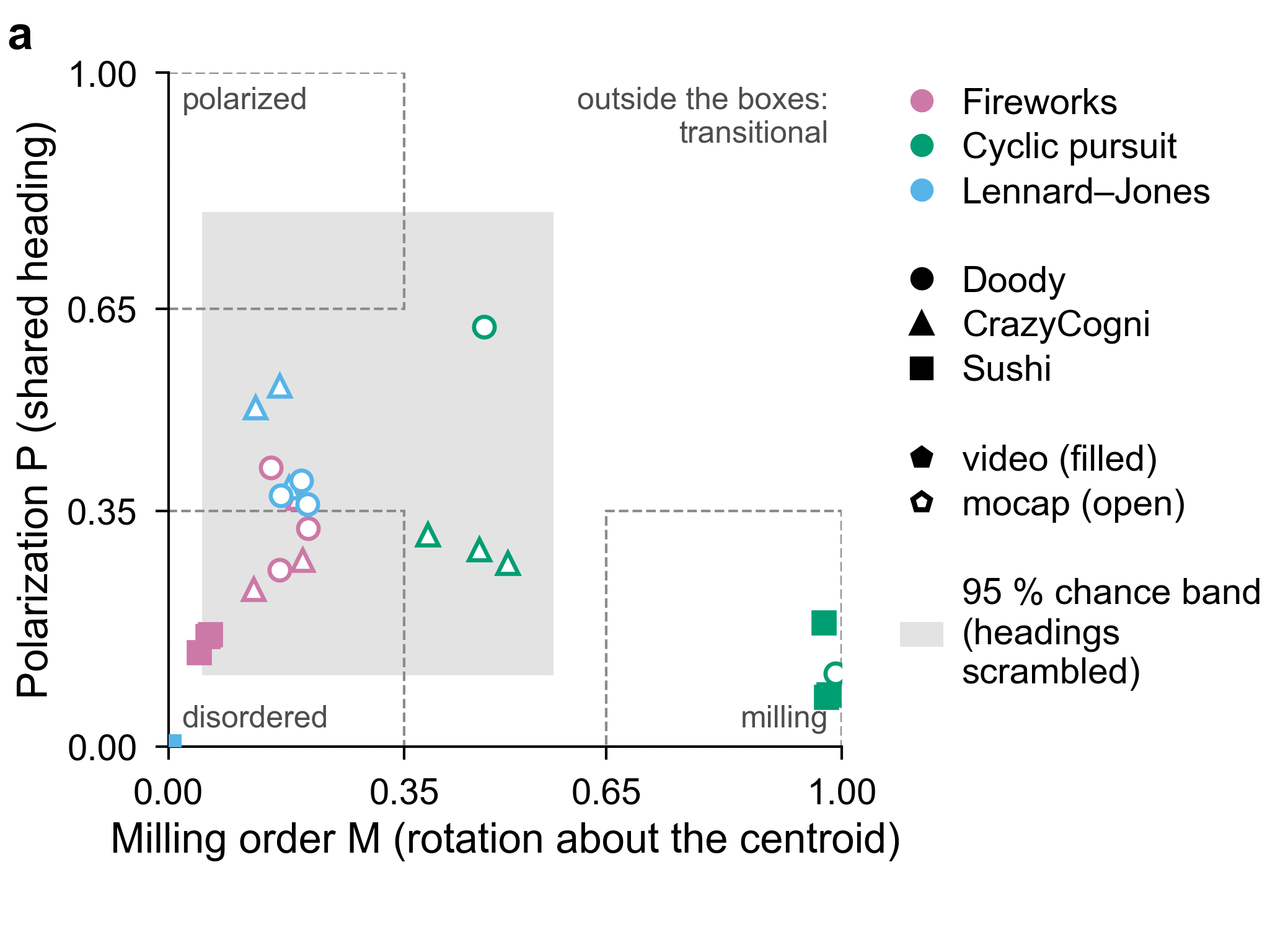}
\caption{polarisation $P$ (shared heading; median over time of $\lVert\langle\hat{\mathbf v}_i\rangle\rVert$) against milling order $M$ (rotation about the centroid; median over time of $|\langle(\hat{\mathbf r}_i\times\hat{\mathbf v}_i)_z\rangle|$), one point per take. Dashed boxes indicate the fish-school states of Tunstr{\o}m et al.~\cite{tunstrom2013collective}: disordered ($P,M<0.35$), milling ($P<0.35$, $M>0.65$), and polarised ($P>0.65$, $M<0.35$). These thresholds were defined for groups of 30--300 fish; with four to six robots, chance alone gives $P\approx0.4$. Grey regions show the union of the 95\% null intervals across takes, each estimated from 200 randomised copies in which every robot's velocity series was independently time-shifted and rotated while positions were preserved.}
\label{fig:s21order}
\end{figure*}

Standard polarisation and milling measures did not capture all of the collective organisation present in the realised behaviours (Fig.~\ref{fig:s21order}). For Fireworks, neither $P$ nor $M$ exceeded its take-specific 95\% null interval, consistent with a behaviour organised by radial expansion and contraction rather than shared heading or rotation. The radial index captured this structure: $\psi$ lay above its null interval for eight of the nine Fireworks takes and below it for all eight Pursuit takes. Lennard--Jones occupied a different intermediate regime: polarisation exceeded the null interval on four of the six room-scale takes but remained below 0.54, indicating only weak-to-moderate alignment rather than coherent collective translation. We could not compute $P$ or $M$ for Lennard--Jones on the Sushis. The robots were mostly still between short corrections that held their spacing, and at 78--83\% of time steps no robot moved fast enough to have a heading, so the averages over robots that define $P$ and $M$ were undefined for most of each take (Table~\ref{tab:realised}). Together, these results show that alignment and milling alone are insufficient to characterise behaviours whose organisation is primarily radial.

At the level of defining collective organisation, Fireworks and Pursuit
transferred successfully across all three embodiments. Their physical
realisation nevertheless differed substantially: Fireworks cycled every 7.6,
12.0, or 26.9~s and varied in directional differentiation; Pursuit produced
circular trajectories ranging from centimetres to metres in scale; and
Lennard--Jones ranged from a slowly drifting lattice to a continuously moving
cloud or a near-stationary formation. These differences arose before human
interpretation, through embodiment-specific dynamics and deployment parameters
that were not normalised across platforms. The realised stimulus therefore
differed in spatial scale, geometry, directional structure, and characteristic
timing, even when the defining collective organisation was preserved. The
present measurements do not establish that these differences caused the audience
responses reported later, since the embodiments also differed in morphology,
covering, staging, and sound. They do establish that the three swarms did not
present equivalent physical stimuli. Before examining how audiences interpreted
them in Sections~(\ref{sec:audience_perception_quant}
and~\ref{sec:audience_perception_qual}), Section~\ref{sec:localisation_results}
characterises the localisation uncertainty under which these collective motions
were realised.

\subsection{Localisation performance during expressive behaviours}\label{sec:localisation_results}

Localisation uncertainty constitutes one source of the reality gap because the
collective behaviours are generated from estimated rather than ground-truth
robot positions. For the Doodies and CrazyCognies, we therefore compared onboard
localisation with motion capture during the recorded expressive behaviours to
characterise the magnitude and spatial distribution of the uncertainty under
which the collective motion was realised. The UWB architecture and its overall
localisation performance were characterised previously~\cite{imran2026swarm};
here, the analysis focuses on whether localisation uncertainty varied
systematically across the performance space or behavioural conditions in which
the collective motion was generated.

\begin{figure*}[!h]
\centering
\includegraphics[width=0.95\textwidth]{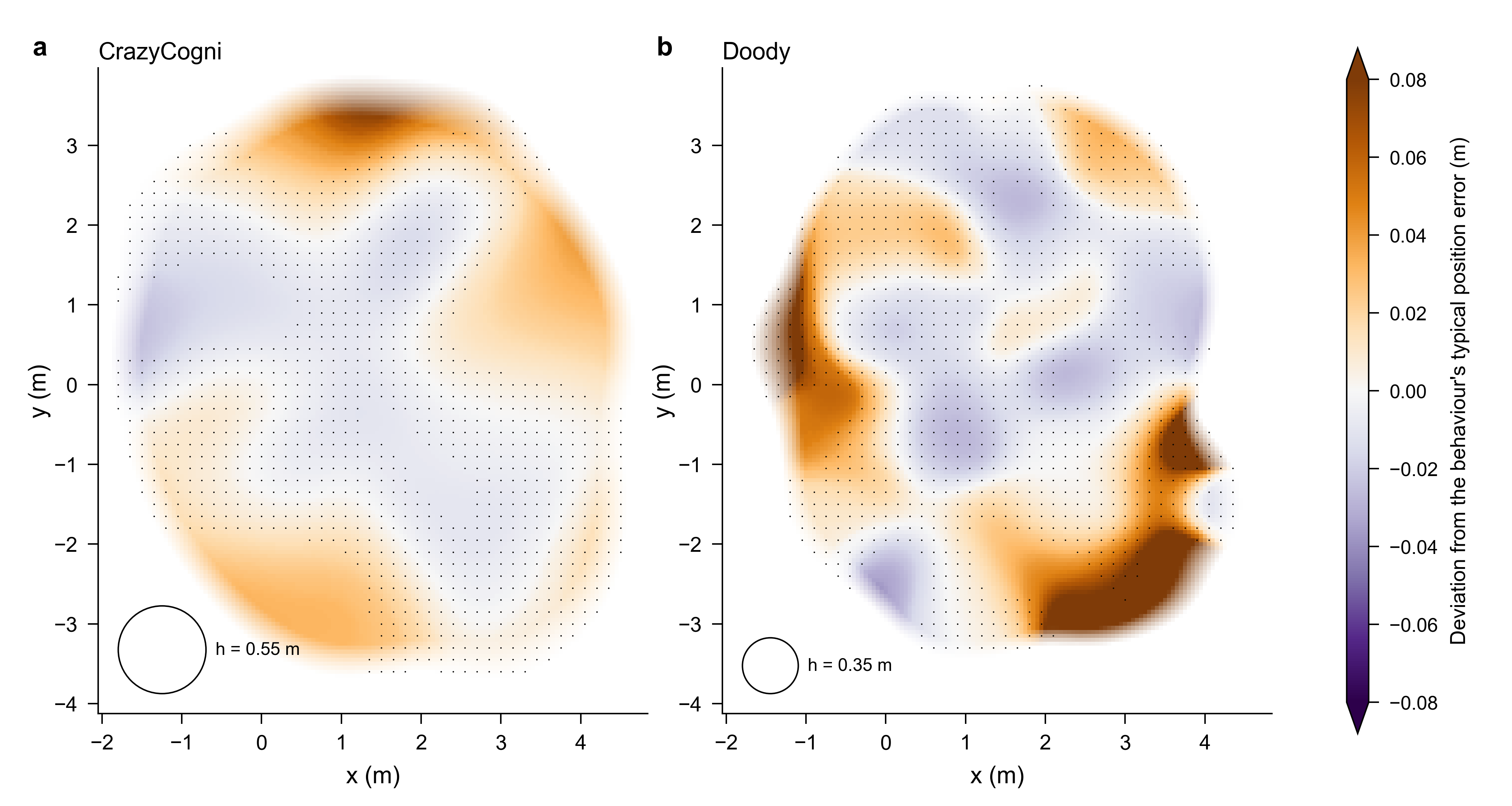}
\caption{Local median position error relative to each behaviour's own median, pooled across behaviours, for the aerial (left) and ground (right) platforms. Stippling marks locations where the local value cannot be distinguished from the platform's typical error, covering 68\% of the aerial field and 83\% of the ground field. Circles indicate the smoothing bandwidth, which differs between platforms because of the number of samples (72,184 ground rows versus 26,120 aerial rows). Smoothing parameters and diagnostics are given in Supplementary Note~2}
\label{fig:localisation1}
\end{figure*}

Across the nine recorded takes per instrumented platform, median horizontal localisation error remained relatively stable across behaviours: 0.077--0.078~m for the CrazyCognies and 0.107--0.154~m for the Doodies (Table~\ref{tab:localisation}). Larger excursions beyond 0.5~m were infrequent, occurring in 0.2--3.0\% of aerial samples and 2.1--6.0\% of ground samples. In total, 41 CrazyCogni robot-takes and 43 Doody robot-takes satisfied the correspondence criteria for comparison with motion capture (Supplementary note~2).

\begin{table}[h!]
\centering
\caption{\textbf{Localisation uncertainty during expressive behaviours.}\label{tab:localisationUncertainty}
Horizontal position error of the onboard estimate relative to motion capture,
by platform and behaviour. Nine takes were recorded per platform, three per
behaviour. $n$ gives the number of robot-takes passing the correspondence criterion. Position error
is reported as median [Q1, Q3]. The final column gives the fraction of
synchronised samples with error exceeding 0.5~m.}
\label{tab:localisation}
\begin{tabular}{llccc}
\toprule
Platform & Behaviour & $n$ & Horizontal error (m) & Samples $>0.5$~m \\
         &           &     & median [Q1, Q3]      & (\%) \\
\midrule
CrazyCognies
    & Fireworks      & 13 & 0.077 [0.049, 0.115] & 1.1 \\
    & Cyclic pursuit & 15 & 0.077 [0.050, 0.111] & 0.2 \\
    & Lennard--Jones & 13 & 0.078 [0.048, 0.111] & 3.0 \\
\midrule
Doodies
    & Fireworks      & 14 & 0.128 [0.081, 0.224] & 6.0 \\
    & Cyclic pursuit & 14 & 0.107 [0.065, 0.155] & 2.1 \\
    & Lennard--Jones & 15 & 0.154 [0.098, 0.224] & 3.1 \\
\bottomrule
\end{tabular}
\end{table}

Previous characterisation of the ground platform attributed larger localisation
errors near arena boundaries to non-line-of-sight conditions and unfavourable
anchor geometry~\cite{imran2026swarm}. This account predicts both a greater
frequency of large errors outside the region occupied during Pursuit and a
systematic change in error with distance from the workspace boundary. Neither
effect was supported. The fraction of samples exceeding 0.5~m was 1.41\% outside
the Pursuit region and 0.92\% inside, with the confidence interval on the
difference including zero. Likewise, the slope of localisation error with
distance from the boundary was \(+0.006\)~m per metre for the CrazyCognies (95\%
CI: \(-0.062\) to \(+0.074\)) and \(-0.016\)~m per metre for the Doodies (95\%
CI: \(-0.065\) to \(+0.034\)). The ground estimate had the direction predicted
by a boundary effect, whereas the aerial estimate had the opposite sign, but
both confidence intervals included zero. The spatial fields show the same
absence of a strong boundary effect (Fig.~\ref{fig:localisation1}): 68\% of the
aerial field and 83\% of the ground field were indistinguishable from each
platform's typical error. The remaining regions differed by only a few
centimetres and did not form a consistent boundary-related pattern.

These measurements characterise the localisation uncertainty under which the
expressive collective motion was realised, rather than providing a benchmark
between platforms, which used different sensing configurations and ground-truth
instrumentation. Typical horizontal errors were on the order of ten centimetres
and were not systematically concentrated near the workspace boundaries. The
collective behaviours therefore remained physically coherent despite
localisation uncertainty of this magnitude. Having characterised both the realised collective dynamics and one source of deployment uncertainty, Sections~\ref{sec:audience_perception_quant}
and~\ref{sec:audience_perception_qual} examine how these embodied collectives
were interpreted by the audience.

\subsection{Quantitative audience perception results}
\label{sec:audience_perception_quant}

Audience responses were obtained from independent cohorts following the three
performances included in the quantitative analysis ($n=13$, 12, and 29,
respectively). The questionnaire assessed meaning attribution, affective
response, appreciation of form, comfort, impressiveness, and perceived
influences of staging and appearance rather than direct inference of a
predefined task goal. Ratings were analysed using cumulative-link mixed models
(Methods), with complete results for all 11 items provided in Supplementary Table~A.

Across performances, the clearest change concerned \emph{appreciation of the
robots' form}, for which the omnibus performance effect remained significant
after correction across the 11 questionnaire dimensions
($\chi^2_{2}=12.68$, $p_{\mathrm{Holm}}=0.019$). Ratings were higher in the
third than in the first performance across embodiments (marginal cumulative
OR $=11.53$, 95\% CI $[3.00,44.34]$, post-hoc
$p_{\mathrm{Holm}}=0.001$); descriptively, median ratings increased from 3 to
4 for the CrazyCognies, from 3 to 5 for the Sushis, and from 2 to 4 for the
Doodies. Several other dimensions, including emotional response,
impressiveness, sympathy, and destabilization, also varied descriptively
across performances, but their omnibus performance effects did not remain
significant after correction across questionnaire items
($p_{\mathrm{Holm}}\geq0.091$). No performance-by-embodiment interaction
survived correction for any of the 11 dimensions, indicating that the data do
not provide strong evidence that the evolution of ratings across performances
differed systematically between robotic systems.

A clearer and more selective pattern emerged across robotic embodiments. Significant omnibus embodiment effects were found for sympathy
($\chi^2_{2}=18.72$, $p_{\mathrm{Holm}}<0.001$), hostility
($\chi^2_{2}=18.46$, $p_{\mathrm{Holm}}<0.001$), the ability to attribute
meaning to movement ($\chi^2_{2}=14.84$, $p_{\mathrm{Holm}}=0.005$),
appreciation of form ($\chi^2_{2}=19.17$, $p_{\mathrm{Holm}}<0.001$),
comfort in the robots' presence ($\chi^2_{2}=12.64$,
$p_{\mathrm{Holm}}=0.013$), and the perceived influence of spatial staging
($\chi^2_{2}=11.78$, $p_{\mathrm{Holm}}=0.017$).

Pairwise contrasts localised most of these embodiment differences to the Sushis. Participants had greater odds of rating the Sushis as
sympathetic than either the CrazyCognies (OR $=3.43$, 95\% CI
$[1.58,7.45]$, $p_{\mathrm{Holm}}=0.004$) or the Doodies
(OR $=5.94$, 95\% CI $[2.60,13.60]$, $p_{\mathrm{Holm}}<0.001$).
Their movements were also more readily attributed meaning than those of the
CrazyCognies (OR $=4.14$, 95\% CI $[1.87,9.17]$,
$p_{\mathrm{Holm}}=0.001$) or Doodies (OR $=3.85$, 95\% CI
$[1.70,8.72]$, $p_{\mathrm{Holm}}=0.002$). Their form was more positively
evaluated (Sushis vs.\ CrazyCognies: OR $=3.71$, 95\% CI $[1.72,8.03]$,
$p_{\mathrm{Holm}}=0.002$; Sushis vs.\ Doodies: OR $=5.78$, 95\% CI
$[2.54,13.17]$, $p_{\mathrm{Holm}}<0.001$), and participants reported
greater comfort in their presence (OR $=2.51$, 95\% CI $[1.15,5.49]$,
$p_{\mathrm{Holm}}=0.041$; and OR $=4.26$, 95\% CI $[1.89,9.58]$,
$p_{\mathrm{Holm}}=0.001$, respectively). Conversely, the Sushis elicited
substantially lower hostility ratings than the CrazyCognies (OR $=0.15$,
95\% CI $[0.05,0.42]$, $p_{\mathrm{Holm}}<0.001$) and the Doodies
(OR $=0.11$, 95\% CI $[0.04,0.30]$, $p_{\mathrm{Holm}}<0.001$).

Spatial staging also differentiated the Sushis from the other embodiments.
Participants were more likely to report that staging influenced their
perception of the Sushis than of the CrazyCognies (OR $=4.87$, 95\% CI
$[1.97,12.06]$, $p_{\mathrm{Holm}}=0.002$) or Doodies (OR $=2.84$,
95\% CI $[1.16,7.00]$, $p_{\mathrm{Holm}}=0.046$). In contrast, the
influence attributed to appearance or costume did not differ significantly
between embodiments after correction ($p_{\mathrm{Holm}}=0.285$). The
Sushis advantage was therefore not a general elevation of all questionnaire
ratings: omnibus embodiment effects were not significant for impressiveness,
emotional response, indifference, destabilization, or appearance influence
after correction (all $p_{\mathrm{Holm}}\geq0.285$). Rather, their
perceptual profile was specifically characterised by greater sympathy, comfort, appreciation of form, and meaning attribution, together with lower perceived hostility.

A notable local pattern concerned the Doodies during the second performance,
where perceived hostility reached a median of 3 [1.75--4], compared with
0 [0--1] in the first and 1 [0--3] in the third performance. The model
contrast for the Doodies alone indicated substantially higher odds of
hostility in performance~2 than performance~1 (OR $=22.65$, 95\% CI
$[2.23,229.61]$), but this contrast did not remain significant after
correction across the within-item post-hoc comparisons
($p_{\mathrm{Holm}}=0.075$). Moreover, the overall
performance-by-embodiment interaction for hostility was not significant
($p_{\mathrm{Holm}}=1.000$). The performance-2 hostility peak is therefore
retained as a descriptive feature of the audience response rather than
evidence for a statistically supported embodiment-specific trajectory. Its
relationship to the scale, staging, costume, and proximity conditions of that
performance is considered later in the Discussion. Descriptive audience-perception profiles across the three embodiments are illustrated in Supplementary Fig.~A.

\subsection{Qualitative patterns in audience interpretation}\label{sec:audience_perception_qual}

Open-ended responses revealed distinct ways in which audiences interpreted the
three robotic embodiments in their performance settings. The Sushis were
predominantly associated with biological and collective imagery. Participants
referred to insects, ants, cells, microorganisms, fish schools, and other small
organisms, while descriptions of what the robots appeared to communicate
frequently invoked collaboration, organisation, mutual assistance, community,
and shared objectives. These readings were especially prominent in the third
performance, where participants frequently attributed coordinated purpose and
social organisation to the group. The qualitative responses therefore portray
the Sushis as a collective whose motion was readily interpreted in relational
and group-level terms.

%Audience comments also suggest that changes in collective state were perceptually salient. During the iterative laboratories, spectators commented explicitly on changes in speed and rhythm and on the transformation of apparently random motion into more organised patterns. Such observations motivated the later emphasis on transitions between aggregation, diffusion, pursuit, attraction, and repulsion as part of the behaviour-composition process. They do not establish transitions as a unique causal source of legibility, but they identify them as a candidate perceptual feature warranting more controlled study.

The CrazyCognies elicited more heterogeneous interpretations. Participants
associated their motion with flying or floating organisms such as insects,
birds, and jellyfish, but also with technological and occasionally military
imagery. Participants attributed intentions ranging from curiosity, observation,
and vulnerability to autonomy and threat. Rather than converging towards a single interpretation, the CrazyCognies supported several simultaneous interpretations of collective agency, combining organic, technological and affective associations.

The Doodies produced the most ambivalent qualitative responses. Their movements
generated anthropomorphic and zoomorphic associations, including dancers,
humans, swans, birds, giraffes, and other large animals. Positive readings
emphasized grace, elegance, coordination, and solidarity. At the same time,
participants also attributed power, intrusion, confrontation, and threat to
them. These interpretations were particularly salient during the second
performance, where responses included aggressive machines, territorial control,
and domination of the human performers. During the third performance,
descriptions of graceful and dancer-like movement coexisted with references to
military figures, physical constraint, and confrontation.

Across embodiments, the qualitative responses indicate that greater meaning
attribution was not equivalent to greater positive affect. The Sushis were
predominantly interpreted through affiliative and collective metaphors, the
CrazyCognies through a broader mixture of organic and technological
associations, and the Doodies through a combination of expressive engagement and
tension. Representative performance excerpts are provided in the performance videos online\footnote{\url{https://initrobots.ca/dessaim/\#performances}}. These distinctions complement the quantitative results by showing that
audiences attributed qualitatively different forms of agency, social
organisation, and affect to the three robotic systems.

%\todonotes[inline]{Report qualitative results from open-ended impressions and discussions. Focus on shifts in how audiences described the swarms: from industrial/threatening/noisy readings toward biological, organic, collective, community, hybrid, or empathic readings if supported by data.}

%\todonotes[inline]{Analyze these qualitative results as evidence of the perceptual gap narrowing through behaviour, embodiment, scenography, and mediation. Avoid claiming causality unless the design supports it.}

%\todonotes[inline]{Use short translated excerpts only if consent and copyright/ethics allow. Otherwise summarise themes and provide the coding procedure in Methods.}

%%%%%%%%%%%%%%%%%%%%%%%%%%%%%%%%%%%%%%%%%%%%%%%%%%%%%%%%%%%%%%%%%%%%%%
\section{Discussion}\label{discussion}
%%%%%%%%%%%%%%%%%%%%%%%%%%%%%%%%%%%%%%%%%%%%%%%%%%%%%%%%%%%%%%%%%%%%%%

The DESSAIM results support a sequential view of the reality and perceptual
gaps. A collective behaviour is first realised by a particular robotic
embodiment and only then becomes available for human interpretation. The two
stages are analytically distinct but experimentally coupled: differences
introduced during realisation determine the stimulus on which interpretation
operates. The present results do not show that either gap has been closed;
rather, they identify complementary technical and perceptual quantities through
which both can be observed and acted upon.

% \subsection{Preserving collective organisation does not preserve physical expression}

At the realisation stage, successful transfer of collective organisation did not
imply equivalence of physical expression. Fireworks and Pursuit retained their
defining organisation on all three embodiments, yet their realised
spatiotemporal characteristics differed: directional separation
was markedly compressed on the CrazyCognies, Pursuit occupied trajectories
spanning centimetres to metres, and the Fireworks cycle ranged from 7.6 to
26.9~s. The same decentralised behavioural specification can therefore survive
deployment while producing substantially different observable collective motion.

Fireworks timing illustrates a limit of platform
abstraction. Its transition counter advanced with each platform's execution frequency. On the Doodies
and CrazyCognies, one cycle required essentially the same number of control
steps, yet controller frequencies differed by a factor of 3.57 and measured
periods by 3.55. The behavioural representation was preserved, but the physical
meaning of a controller-step parameter was not. Architectures such as Buzz and PyBuzz can standardise sensing and motion
interfaces without automatically normalising the physical timescale of local
interactions~\cite{pinciroli2016buzz,rosbuzz2020}.

Transfer criteria must also reflect the organisation
produced by a behaviour. Polarisation and milling describe aligned and rotating
collective states~\cite{vicsek1995novel, couzin2002collective,
tunstrom2013collective}, but did not capture Fireworks' radial contraction--expansion
structure, whereas the radial index did. Validation
should therefore establish whether the defining collective organisation is
recoverable from realised motion rather than rely on a single generic measure of
order. Localisation uncertainty was one source of discrepancy, but did not explain the cross-platform differences by itself:
horizontal errors were typically around ten centimetres, showed no
systematic boundary effect, and the behaviours remained physically coherent. The
reality gap of a swarm is consequently better considered at the level of the
resulting interaction process than as an accumulation of independent agent-level
errors.

% \subsection{Collective organisation as a perceptual object}

These differences matter because observers encounter the realised collective motion, not the controller specification. Whereas single-robot legibility commonly concerns inferring an intended goal from motion~\cite{dragan2013legibility,dragan2015effects}, multi-robot studies show that coordination objectives, collective motion patterns, and spatial distributions can themselves provide perceptual information~\cite{capelli2024,kim2021generating,ichihashi2024swarm}. In DESSAIM, audience interpretations involved both individual bodies and their relationships: the Sushis were described through biological and collective metaphors, while the Doodies elicited both individual and group-level readings. Their articulated bodies prompted anthropomorphic and zoomorphic associations, while displacement and regrouping were interpreted as coordinated collective activity. The perceptual object of a swarm may therefore shift between its constituent bodies and their collective organisation rather than be confined to a single channel of apparent intent. Unlike human--swarm transparency research focused mainly on operators using dedicated interfaces~\cite{hepworth2020human,patel2022multi,kolling2015human}, these interpretations arose through direct observation of expressive collective motion.

Transitions between collective states also emerged as a candidate expressive feature at the group level. Aggregation, diffusion, Pursuit, and attraction or repulsion create changes in density, circulation, and spatial organisation, while Fireworks sequences collective regimes. Informal
observations during the research-creation laboratories repeatedly referred to
changes in speed, rhythm, organisation, and apparent randomness. Because these
observations were design inputs rather than formally coded audience data, and
transitions were not manipulated independently, they do not establish that
transitions are more readily interpreted than steady states. They do, however,
motivate transitions as units for future controlled study, consistent with
recent performing-arts approaches to robot-group
movement~\cite{Madill2025PlayingRobots}.

Movement Observation--Analysis provided a practical bridge between perceptual
observations and design intervention. MOA links functional and expressive
dimensions of movement through Ground, Space, and
Dynamics~\cite{harbonnier2021}. In DESSAIM, impressions were reformulated as
observable movement properties that could inform changes in behaviour
parameters, sequencing, embodiment, or staging. MOA therefore served as a shared
observation--description--intervention vocabulary, making perceptual
discrepancies actionable without reducing subjective interpretation to a
numerical movement model.

% \subsection{Embodiment, context, and audience interpretation}

Collective motion was only one component of what audiences encountered: the three systems also differed in scale, morphology, degrees of freedom, sound, coverings, staging, and proximity to performers and spectators. These factors were co-designed rather than independently manipulated, preventing attribution of their separate causal effects. This distinction matters because a physical embodiment influences not only the motion a controller can produce, but also the sensory and spatial cues available to an observer. Embodiment therefore forms part of the realised stimulus from which observers construct an interpretation, consistent with research showing that swarm distribution, embodiment, and collective-motion features affect how robot groups are perceived~\cite{kim2021generating,ichihashi2024swarm}.

The performance environment extended expression beyond the robots: airflow animated scenographic materials, projected shadows altered apparent spatial extent, amplified mechanical sounds provided another sensory channel, and the Doodies' articulated arms added postural variation. These effects were not isolated experimentally; rather, they illustrate how swarm expression can be distributed across robots, their relationships, and the environment they affect. Robot performance and choreography similarly treat body, movement, spatial relationships, and staging as coupled components of expression~\cite{naoko2022,cuan2020,Jochum2016RobotChoreography}.

The questionnaire and qualitative responses further distinguish meaning attribution from affect. Corrected omnibus embodiment effects were found for sympathy, hostility, meaning attribution, appreciation of form, comfort, and perceived influence of spatial staging, with most pairwise differences involving the Sushis. These effects were selective rather than general across questionnaire dimensions, and no performance-by-embodiment interaction survived correction. The absence of other corrected effects should not be taken as evidence of perceptual equivalence, particularly given the small and unequal cohorts. Across performances, only appreciation of form retained a corrected performance effect. The predominantly affiliative interpretations of the Sushis contrasted with the more ambivalent associations elicited by the CrazyCognies and Doodies, including vulnerability, intrusion, and threat. Greater meaning attribution was therefore not equivalent to more positive affect. For human-facing swarms, meaning attribution, affect, aesthetics, comfort, and acceptance should remain distinct dimensions rather than be collapsed into a single notion of legibility.

% \subsection{Public performance as an ecological testbed}

These findings also clarify the methodological value of public performance for examining both gaps together. Performance-oriented HRI combines embodied interaction, movement expertise, and audience observation outside conventional laboratories~\cite{Jochum2016RobotChoreography,Nishiguchi2017Theatrical,StOnge2017RHI}. HRI research more broadly recognises the trade-off between experimental control and ecological validity~\cite{HoffmanZhao2021}, while recent performing-arts workshops have informed movement vocabularies for robot groups~\cite{Madill2025PlayingRobots}. DESSAIM extends these approaches to heterogeneous decentralised swarms observed in repeated public performances.

Repeated performances required the swarms to operate across changing venues, with portable localisation, shared 2.4~GHz wireless communication, variable scenography, and close proximity to performers and spectators. Venue dimensions, lighting, sound, audience viewpoint, and sequence duration shaped both the collective motion realised and what audiences could observe. The simultaneous pressure on deployment and interpretation would be difficult to recreate in an isolated controller or perception benchmark. Although such performances are not equivalent to industrial, agricultural, search-and-rescue, or other field deployments, they provide an ecological setting in which technical realisation and human interpretation must coexist.

Performance thus served as an ecological stress test of the coupling between the two gaps, not a substitute for controlled validation. Common behavioural code remained operational even as characteristic timing and directional structure varied across embodiments, and physically coherent collectives supported different human interpretations. The exploratory audience design and coupled changes in motion, embodiment, and staging limit causal attribution and motivate controlled comparisons of collective perceptual cues (Supplementary note~3). Shared behaviour scripts, simulation configurations, localisation analyses, MOA-informed observation procedures, composition examples, and public performance documentation\footnote{\url{https://initrobots.ca/dessaim/\#performances}} support replication of parts of this workflow and further tests of the two-gap framing in human-facing applications.

%%%%%%%%%%%%%%%%%%%%%%%%%%%%%%%%%%%%%%%%%%%%%%%%%%%%%%%%%%%%%%%%%%%%%%
\section{Methods}\label{sec:methods}
%%%%%%%%%%%%%%%%%%%%%%%%%%%%%%%%%%%%%%%%%%%%%%%%%%%%%%%%%%%%%%%%%%%%%%
The methods are organised around two axes corresponding to the two gaps examined in Section~\ref{sec:results}. The first concerned the simulation-to-deployment reality gap: how collective behaviours specified and tested in software and simulation were realised on heterogeneous physical robots under venue-specific sensing, communication, localisation, and dynamical constraints. The second concerned the perceptual gap: how realised collective motion was composed, observed, staged, and iteratively modified in response to artistic observation and audience interpretation.

The two axes were not intended as procedures that definitively close either gap. They provided complementary means of making each gap observable and actionable during development. Axis 1 maintained a common behavioural representation across embodiments while exposing platform-specific constraints on its physical realisation. Axis 2 treated behaviour, transitions, embodiment, scenography, sound, appearance, performers, and mediation as a coupled expressive design space. These factors were co-designed rather than factorially manipulated; the present study therefore does not estimate their independent causal contributions.

\subsection{Axis 1: The realisation step}

\subsubsection{Robotic platforms}\label{subsec:Platforms}

\begin{figure*}[htp]
    \centering
    \subcaptionbox{Sushi\label{fig:ZooidComponents}}{%
        \includegraphics[height=1.6in]{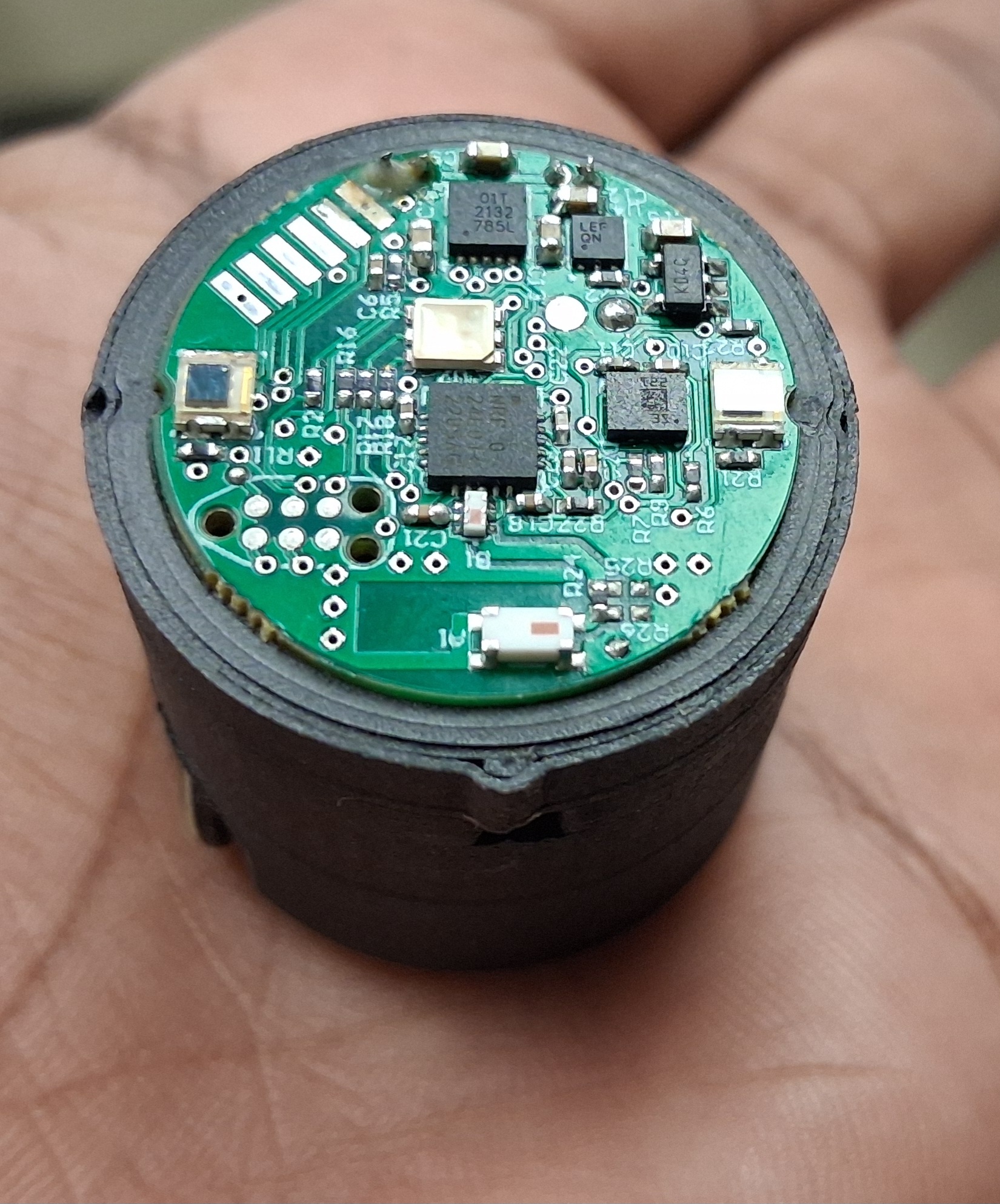}%
    }\quad
    \subcaptionbox{CrazyCognie\label{fig:CogniComponents}}{%
        \includegraphics[height=1.6in]{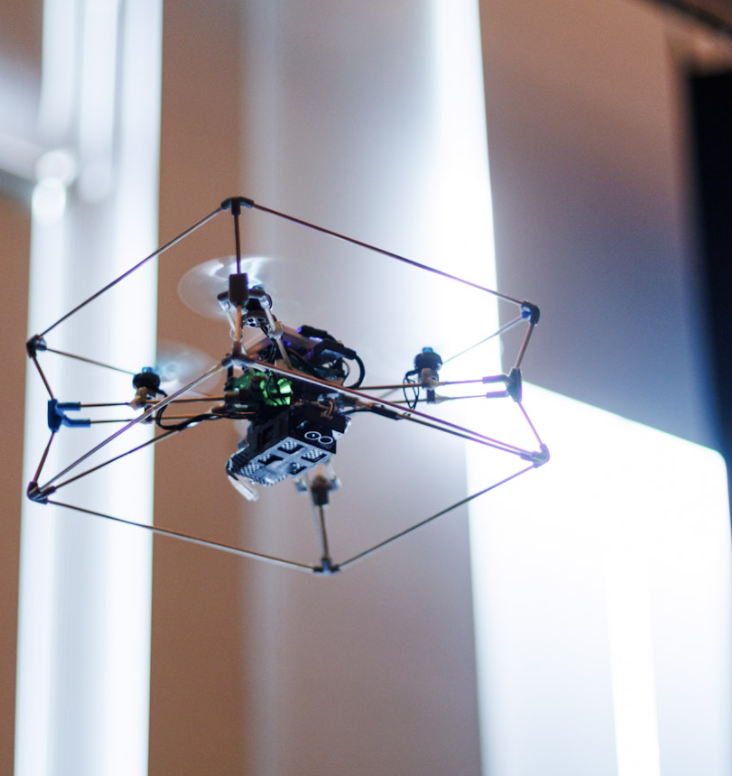}%
    }\quad
    \subcaptionbox{Doody\label{fig:dingonu}}{%
        \includegraphics[height=1.6in]{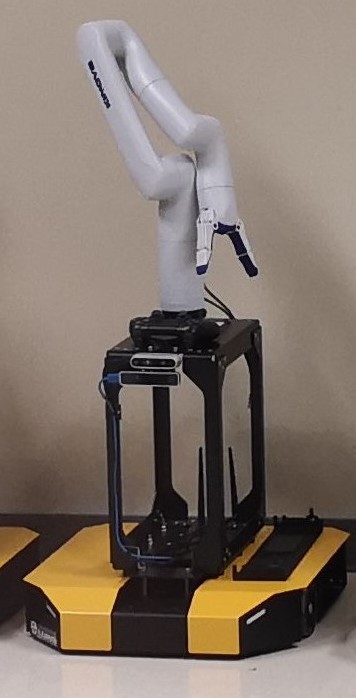}%
    }
    \caption{The three DESSAIM robotic platforms.\label{fig:swarmsComponents}}
\end{figure*}

The realisation step concerns how a common behavioural specification becomes physical collective motion across heterogeneous robotic embodiments. Figure~\ref{fig:swarmsComponents} shows the three robotic platforms used in DESSAIM: 2~cm tabletop Sushis, 30~cm caged aerial CrazyCognies, and human-scale mobile manipulators.

\textbf{Sushis.}
The DESSAIM fleet comprises 24 miniature tabletop robots derived from Stanford's Zooids~\cite{le2016zooids}, of which typically 12 were deployed during performances. Relative to the original platform, the wheel drive was modified with a higher gear ratio, trading maximum speed for finer control at the lower velocities required for expressive collective motion. As shown in Fig.~\ref{fig:ZooidComponents}, each robot uses differential-drive locomotion actuated by two small DC motors. Two photodiodes provide localisation from the projected tracking system, while a wireless module supports communication with the swarm-control infrastructure. Although the behavioural control law is decentralised at the agent level, computation is executed on a ground station because of the limited onboard resources of the Sushis. A single-cell LiPo battery provides approximately 1~h of operation. Their centimetre-scale dimensions and planar workspace support dense collective motion within a compact tabletop environment.

\textbf{CrazyCognies.}
The DESSAIM fleet comprises 12 CrazyCognies (Fig.~\ref{fig:CogniComponents}), of which typically four were deployed during performances. The platform is based on the collision-resistant Cognifly~\cite{de2022being} and incorporates the Crazyflie sensing stack. Each robot executes its decentralised behavioural controller onboard using a Raspberry Pi 3A. A custom software stack based on the Cognifly driver~\cite{Bouteiller2024thecognifly} connects the vehicle's actuation, sensing and pose-estimation components. The surrounding cage enables operation in confined indoor environments and in proximity to performers and audiences. A three-cell LiPo battery provides approximately 15~min of flight. Unlike the ground-based embodiments, the CrazyCognies realise collective motion in three dimensions, adding vertical displacement and aerial dynamics to the shared behavioural repertoire.

\textbf{Doodies.}
The DESSAIM fleet comprises eight human-scale mobile manipulators (Fig.~\ref{fig:dingonu}), of which typically six were deployed during performances. Each robot combines a Clearpath Dingo differential-drive base~\cite{clearpath_dingo_2024} with a Kinova Gen3 Lite manipulator~\cite{kinova_gen3_lite_2024}. Behavioural computation runs onboard on an NVIDIA Xavier, while a Crazyflie Loco deck forms part of the pose-estimation system. The base and manipulator are powered independently by two 12~V lead-acid batteries, providing more than 3~h of continuous operation. At a scale comparable to human performers and audience members, the Doodies combine planar collective locomotion with articulated arm motion, extending the realised behaviour beyond base trajectories alone.

\subsubsection{PyBuzz simulation-to-deployment workflow}\label{subsec:pybuzz}

%\todonotes[inline]{Describe PyBuzz as the toolchain used to move behaviours from simulation/design to physical robots. Include Buzz scripts, Python integration, Cython bridge, Buzz VM, hardware abstraction, adapters, standardised position representation, motion commands, neighbour discovery, and supported simulation environments.}

%\todonotes[inline]{Phrase claims carefully: PyBuzz is the deployment architecture used in DESSAIM and a reusable tool to be released. Do not claim a general-purpose benchmarked framework unless the paper includes such validation.}

Buzz is a swarm-oriented scripting language designed for decentralised multi-robot behaviours~\cite{pinciroli2016buzz}. A Buzz program specifies the local logic executed independently by each agent, allowing collective behaviour to emerge from repeated interactions among neighbouring robots rather than from a centralised trajectory specification. In DESSAIM, we used \textit{PyBuzz}, a Python-based integration layer around the Buzz Virtual Machine (BVM), to preserve this behavioural representation across simulation and heterogeneous physical platforms.

\begin{figure*}[ht!]
    \centering
    \includegraphics[width=0.94\textwidth]{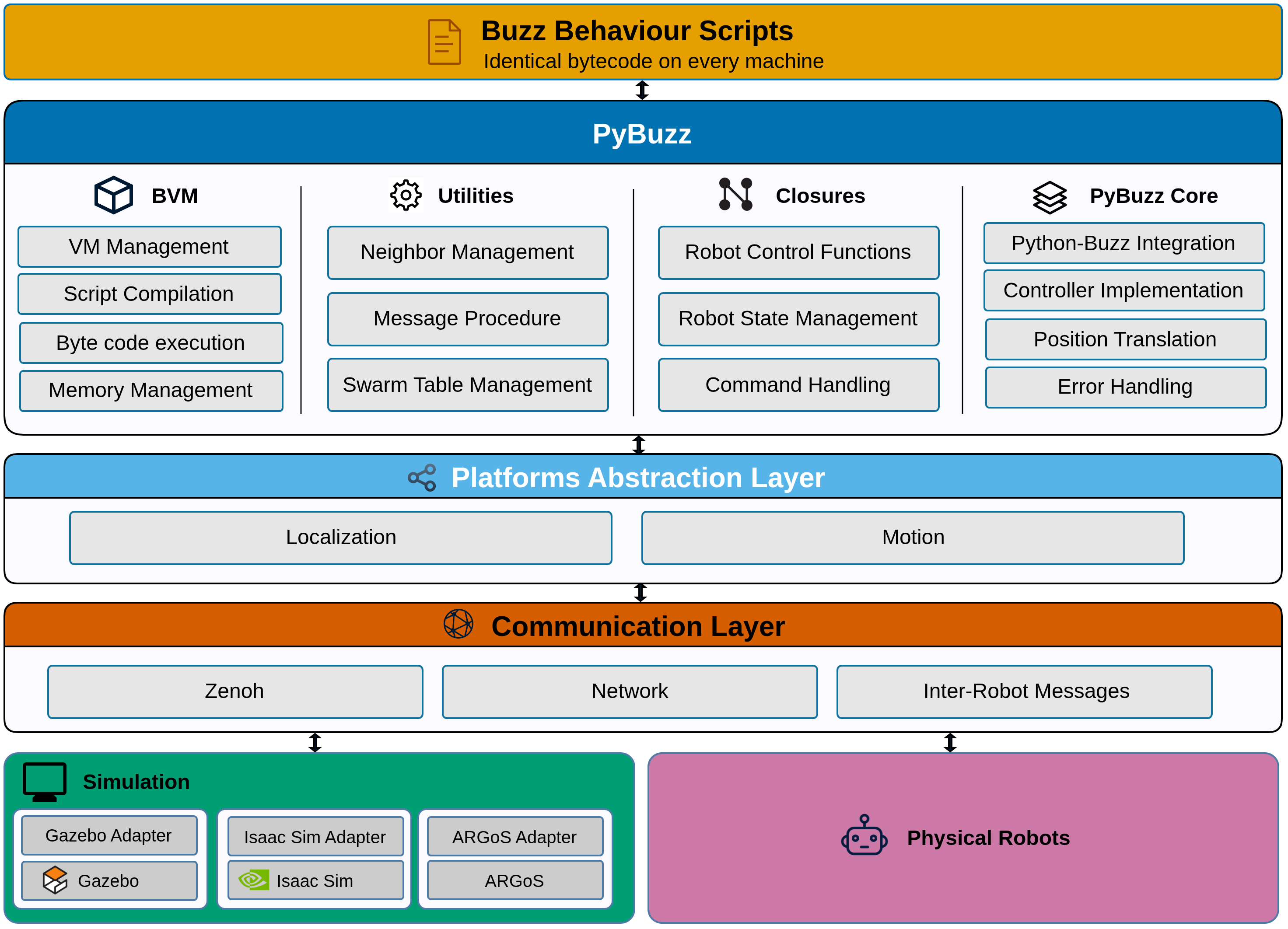}
    \caption{PyBuzz architecture used to preserve a common Buzz behavioural representation across simulation and heterogeneous physical platforms. Each logical robot executes an independent Buzz Virtual Machine (BVM), while platform adapters translate localisation, sensing, inter-robot communication, and motion commands between the behavioural layer and platform-specific interfaces.}
    \label{fig:BuzzArchitecture}
\end{figure*}

PyBuzz connected behaviour design, simulation, and physical deployment through a common execution interface (Fig.~\ref{fig:BuzzArchitecture}). Each logical robot executes its own BVM instance, preserving the decentralised execution model of Buzz. The BVM is sufficiently lightweight to run directly on embedded processors, including microcontrollers of the class used by the Sushis. For the DESSAIM performances, however, the Sushi BVM instances were hosted on a ground station to simplify operation and supervision of the larger tabletop swarm. The CrazyCognies and Doodies executed their behavioural programs onboard.

PyBuzz provides a bidirectional interface between each BVM and the underlying robotic platform. The behavioural program can access the robot's own localisation estimate, the positions of neighbouring robots, messages exchanged between neighbours for behavioural coordination, and additional platform-specific sensor measurements when required. Commands generated by the Buzz program are translated through platform adapters into the motion representation expected by each robot. In the staged behaviours examined in this study, no exteroceptive sensing other than localisation was used; collective coordination relied on robot positions and, where required, messages exchanged between neighbouring agents.

The adapters standardise the quantities required by the swarm program while leaving the underlying sensing, localisation, communication, and actuation architectures platform-specific. Localisation data originating from different systems are converted into a common spatial representation from which neighbour relations can be constructed, while high-level motion commands are mapped to the appropriate platform interface. The same Buzz behavioural scripts could therefore be executed on the Sushis, CrazyCognies, and Doodies despite their substantial differences in scale, dynamics, localisation systems, and computational architecture.

This abstraction does not normalise all physical quantities. In particular, parameters expressed in controller steps retain the same numerical value across platforms while inheriting the execution frequency of the platform on which the BVM runs. PyBuzz therefore preserves the behavioural specification without imposing a common physical time base. This distinction is directly relevant to the Fireworks timing differences reported in Section~\ref{subsec:EmergentBehaviours}.

PyBuzz also interfaces with multiple simulation environments, allowing the same Buzz behavioural representation to be developed and tested before physical deployment. In this work, PyBuzz is treated as enabling infrastructure rather than as a standalone software contribution: it provided the common experimental substrate required to examine how a shared decentralised behavioural specification was physically realised by heterogeneous robotic embodiments.

\subsubsection{Communication infrastructure}\label{sec:comms}

%\todonotes[inline]{Describe Zenoh architecture, publish/subscribe/query abstractions, gossip messages, single-topic strategy, packet overhead, ground stations, launch/switch/stop commands, and procedures for adding/removing robots. This supports reproducibility of the deployment, not the Results claim by itself.}

Inter-robot communication provides the state exchange required by the decentralised behaviours. In the DESSAIM deployments, robots within each swarm operated with a fully connected logical communication topology: each agent could receive the state information and coordination messages broadcast by every other active agent. Although Buzz and PyBuzz can support neighbourhood-limited communication and other network topologies, these were not used in the performances reported here.

For the Sushis, individual BVM instances ran on a ground station, which transmitted position commands to the robots through a 2.4~GHz radio interface. For the CrazyCognies and Doodies, inter-robot exchange used Zenoh~\cite{zenoh}, a lightweight publish/subscribe protocol. Each robot periodically broadcast its local state and behaviour-coordination messages through a shared gossip topic. The common topic allowed robots to enter or leave the active swarm without changing the behavioural interface or creating robot-specific communication channels.

All three systems therefore relied on 2.4~GHz wireless communication during public performances. These deployments took place in audience-filled indoor environments, where human bodies could obstruct radio links while numerous personal wireless devices operated in the same frequency band. The expressive behaviours remained operational under these conditions across the performances reported here. Communication latency, packet loss, and link quality were not instrumented systematically, however, so these deployments are reported as operating conditions rather than as a quantitative assessment of network robustness.

\subsubsection{Localisation and calibration}\label{localisation}

%\todonotes[inline]{Describe localisation methods for each platform: projected photodiode tracking for Sushis; UWB TDOA3 and filtering for CrazyCognies; UWB plus wheel odometry/IMU Kalman filtering for Doodies. Include anchor setup, calibration process inspired by land surveying, cost and setup time if retained, and limitations.}

The collective behaviours operate on estimated rather than ground-truth robot
positions, making localisation uncertainty one potential source of discrepancy between the specified behaviour and its physical realisation. Because the DESSAIM platforms operated at different spatial scales and under different deployment constraints, each embodiment used a distinct localisation system while exposing a common position representation to the behavioural layer.

For the tabletop Sushis, localisation was provided by the projector-based tracking system inherited from the Zooids platform. Gray-coded patterns were projected onto the workspace by a 3000~Hz projector (DLP LightCrafter 4500) and detected and decoded by photodiodes mounted on each robot, providing real-time position estimates over the tabletop workspace with an accuracy of approximately 1~cm~\cite{zhao2017robotic}. This system was used directly during DESSAIM and was not independently benchmarked against motion capture in the present study.

The CrazyCognies relied on an Ultra-Wideband (UWB) localisation infrastructure for global horizontal position estimation. Preconfigured anchors were distributed around the performance space, while each aerial robot carried a UWB tag. The system operated in TDOA3 mode~\cite{zhao2021learning}, using differences in signal arrival time at synchronised anchors to estimate robot position. The resulting measurements were filtered to obtain the horizontal pose $(x,y)$ used by the behavioural controller. Altitude was estimated independently using a TFmini lidar, which proved more reliable than UWB-based vertical estimates in the performance environments.

The accuracy of the UWB system depends on anchor placement and calibration. For each deployment, the anchor network was calibrated using the land-surveying-inspired procedure developed previously by our group\footnote{\url{https://git.initrobots.ca/swarmui/buzz_unified}}. The procedure enabled the UWB infrastructure to be installed and calibrated in less than 4~h for a new performance space. The same UWB infrastructure was used by the Doodies, for which the global position estimates were additionally fused with wheel odometry and inertial measurements through a Kalman filter.

Localisation accuracy was evaluated against external motion capture for the CrazyCognies and Doodies during the recorded expressive behaviours. A \emph{robot-take} denotes the trajectory of one robot during one recorded take. For the CrazyCognies, the onboard record was the pose estimated by each robot and broadcast to its neighbours; for the Doodies, it was the fused UWB--odometry--IMU estimate. The motion-capture trajectory was treated as the external reference representing the physically realised motion, whereas the onboard trajectory represented the spatial state available to the robot during behaviour execution.

Onboard and motion-capture trajectories were processed to remove isolated spikes and implausible jumps, resampled to 10~Hz and temporally aligned using one time shift per take. Robot-takes were retained only when the onboard trajectory could be placed in correspondence with its own motion-capture body, using a median-distance criterion of 0.5~m (Supplementary note~2). A single rigid transform per platform was then fitted across all retained takes to express onboard positions in the motion-capture frame without removing take-specific localisation bias.

Horizontal localisation error, $e_{xy}$, was computed as the Euclidean distance between synchronised onboard and motion-capture positions within each take's active window. For each platform and behaviour, error was summarised by its median and quartiles and by the fraction of samples exceeding 0.5~m (Table~\ref{tab:localisationUncertainty}). Behavioural comparisons were performed on robot-take summaries, with uncertainty clustered at the robot-take level.

Spatial structure in localisation error was examined in two complementary ways. First, error was tested as a function of distance from the workspace boundary and, for the ground platform, compared inside and outside the region occupied during Pursuit (Supplementary Note~2). Second, local error fields were estimated from behaviour-centred errors using kernel-weighted local medians. Uncertainty in the local field was estimated by bootstrap resampling at the robot-take level. Kernel bandwidths and visualisation thresholds are reported in Sec.~\ref{sec:localisation_results}.

\subsubsection{Analysis of realised collective motion}\label{subsec:analysis}

The trajectory analysis was designed to characterise properties of the collective motion that remain observable across embodiments: spatial scale, overall speed, radial or tangential organisation, alignment, and circulation. Following the trajectory preprocessing described in Section~\ref{localisation}, velocities were estimated from positional differences over $\pm0.1$~s and smoothed using a 0.5~s moving average. Analyses were restricted to time steps for which all robots in a take were tracked.

At each time step, we computed the swarm centroid $\mathbf c(t)$ and each robot's outward radial direction,
$\hat{\mathbf r}_i=(\mathbf p_i-\mathbf c)/
\lVert\mathbf p_i-\mathbf c\rVert$.
Velocity was decomposed into components parallel to this direction (radial) and perpendicular to it (tangential). The spatial extent of the swarm was quantified by the radius of gyration,
\[
R_g(t)=
\left(
N^{-1}\sum_i
\lVert\mathbf p_i-\mathbf c\rVert^2
\right)^{1/2},
\]
which equals the orbit radius for robots uniformly distributed around a circle. Each take was summarised by the median $R_g$, speed, and absolute radial and tangential speeds over robots and time. Takes whose median speed fell below the platform-specific activity threshold were marked as below the activity gate.

Directional organisation was quantified from the angle $\alpha_i$ between each robot's heading $\hat{\mathbf v}_i$ and its outward radial direction $\hat{\mathbf r}_i$. A heading was defined only when speed exceeded a platform-specific floor chosen to exclude directions dominated by tracking noise: 0.02~m\,s$^{-1}$ for motion capture, 0.037~m\,s$^{-1}$ for video, and
0.18~m\,s$^{-1}$ for the CrazyCogni onboard estimates. We then defined the radial index

\begin{equation}
\psi=\langle\cos 2\alpha_i\rangle,\qquad \psi\in[-1,1],
\label{eq:psi}
\end{equation}

\noindent averaged over all robot samples with a defined heading. Values near $+1$ indicate predominantly radial motion, values near $-1$ predominantly tangential motion, and values near zero no preferred radial--tangential axis. The doubled angle makes the measure insensitive to direction along the radial axis, so inward and outward phases of Fireworks contribute with the same sign rather than cancelling.

For comparison with established descriptors of collective order, we also computed polarisation
$P(t)=\lVert N^{-1}\sum_i\hat{\mathbf v}_i\rVert$
and milling order
$M(t)=\lvert N^{-1}\sum_i
(\hat{\mathbf r}_i\times\hat{\mathbf v}_i)_z\rvert$,
with robots below the heading threshold contributing zero. Each take was summarised by the median value over time. Polarisation and milling are established measures for distinguishing aligned, rotating, and disordered collective states ~\cite{vicsek1995novel,couzin2002collective,tunstrom2013collective}. Because their finite-group distributions depend on the number of agents and on the
recorded trajectories, $P$, $M$, and $\psi$ were also compared with a take-specific null distribution generated from 200 randomised copies. For each copy, individual velocity series were independently time-shifted and rotated while the recorded positions were preserved. The central 95\% interval of the resulting distribution was used as the take-specific null interval.

Two low-motion cases required additional treatment. In the Sushi Lennard--Jones takes, no robot exceeded the heading threshold during 78--83\,\% of time steps; the resulting polarisation and milling values are therefore reported as undefined rather than as zeros produced by the thresholding procedure. For the CrazyCogni onboard estimates, the heading floor was close to the swarm's median speed, making $P$ and $M$ strongly dependent on thresholding; these measures are therefore not reported for that layer. The corresponding $\psi$ values are retained but should be interpreted as biased toward zero.

For the cross-platform comparison, we considered the defining collective organisation of a behaviour to have transferred when the organisation specified in
Table~\ref{tab:behaviors} was recoverable from the externally observed trajectories using the corresponding descriptors above. This criterion concerns preservation of the defining collective organisation, rather than equality of spatial scale, speed, timing, or directional structure across embodiments.

Fireworks periods were estimated from the radius-of-gyration signal using a two-stage least-squares procedure: an initial sinusoidal fit followed by refinement with a three-harmonic waveform, both including a quadratic trend. The estimates were checked against autocorrelation and spectral methods. Periods were reported only when the valid observation span covered at least three estimated cycles, and sensitivity to detrending was assessed using moving-average windows of 12–60 s, excluding windows shorter than the estimated period, eligibility takes and sensitivity results are given in Supplementary Note~1.

%Fireworks timing was analysed separately from its directional structure. The period was estimated from the radius-of-gyration signal $R_g(t)$ by least-squares fitting of a sinusoidal waveform and independently checked using autocorrelation. A period was reported only if the analysed record spanned at least three cycles. To check that the period did not depend on how slow drift was handled, we repeated the estimate after first subtracting moving averages of 12--60~s, skipping any window shorter than the period, which would remove part of the cycle itself.

\subsubsection{Behaviour library and deployment procedure}\label{sec:library}

%\todonotes[inline]{Define the canonical and expressive swarm behaviours used in the project. For each behaviour, specify inputs, tunable parameters, expected collective pattern, platforms deployed, and performance constraints. Keep equations only where necessary or already present.}

The twelve behaviours used in DESSAIM formed a compositional vocabulary built from a small set of decentralised control primitives. Table~\ref{tab:behaviors} summarises their deployment, specified collective organisation, and intended compositional function. The library is generated from two underlying motion-control kernels, combined with a temporal composition mechanism for sequencing and synchronising collective states.

\begin{table*}[!h]
\centering
\small
\caption{Expressive behaviour vocabulary used in DESSAIM. S: Sushis;
C: CrazyCognies; D: Doodies.}
\label{tab:behaviors}
\begin{tabular}{p{0.15\linewidth} p{0.10\linewidth} p{0.34\linewidth} p{0.31\linewidth}}
\toprule
\textbf{Behaviour} &
\textbf{Platforms} &
\textbf{Designed collective organisation} &
\textbf{Compositional function} \\
\midrule

Aggregation &
S &
Convergence toward the local centroid &
Contraction; increasing density \\

Attraction point &
S &
Convergence toward an external reference &
Directed collective displacement \\

Diffusion &
S &
Motion away from the local centroid &
Expansion; decreasing density \\

Repulsion point &
S &
Motion away from an external reference &
Directed collective avoidance \\

Fireworks &
S, C, D &
Cyclic aggregation--diffusion &
Rhythmic contraction--expansion \\

Lennard--Jones &
S, C, D &
Regulation of inter-agent spacing &
Cohesive, deformable spatial structure \\

Flocking &
S &
Coherent displacement through local interactions &
Shared heading and collective translation \\

Pursuit &
S, C, D &
Convergence toward sustained circular motion &
Transition from disorder to circulation \\

Pursuit + attraction &
S &
Pursuit biased by an external reference &
Spatially displaced or deformed circulation \\

Fireworks + Pursuit &
S &
Simultaneous execution by two subgroups &
Juxtaposition of collective regimes \\

Fireworks Spiral &
S &
Aggregation $\rightarrow$ Pursuit $\rightarrow$ Diffusion $\rightarrow$ Aggregation &
Contraction--circulation--expansion sequence \\

Fireworks Curve &
S &
Aggregation $\rightarrow$ Diffusion $\rightarrow$ Pursuit $\rightarrow$ Aggregation &
Contraction--expansion--circulation sequence \\

\bottomrule
\end{tabular}
\end{table*}

Each behaviour is implemented as a decentralised controller executed independently for each logical agent. At every control step, the Buzz program receives the robot's own position and the positions of the other active robots through the common PyBuzz interface described in Sections~\ref{subsec:pybuzz} and~\ref{localisation}. For neighbour $i$, position $\mathbf{p}_i$ relative to the robot position $\mathbf{p}$ is expressed as range $d_i$ and bearing $\theta_i$,

\begin{equation}
d_i = \lVert \mathbf{p}_i - \mathbf{p} \rVert,
\qquad
\theta_i = \operatorname{atan2}\!\left( y_i-y,\;x_i-x \right),
\label{eq:rab}
\end{equation}

from which the controller computes a desired motion command. PyBuzz then maps this command to the corresponding platform interface. The behavioural scripts are therefore shared across the Sushis, CrazyCognies, and Doodies, while numerical parameters and the low-level realisation of the command remain platform-specific. The staged behaviours analysed here used localisation as their only exteroceptive input, together with inter-agent messages where required for behavioural coordination. The controller formulations follow those used previously in our multi-platform experiments ~\cite{imran2026swarm} and are summarised here in terms of the parameters relevant to DESSAIM.

\paragraph{Potential-field motion}
Most behaviours are generated from a weighted sum of neighbour and external-point displacements,

\begin{equation}
\mathbf{u} \;=\; \sum_{j \,\in\, \mathcal{N} \cup \mathcal{Q}}
s_j \, f(d_j)\,\bigl(\mathbf{p}_j - \mathbf{p}\bigr),
\label{eq:potential}
\end{equation}

where $\mathcal{N}$ is the set of currently known neighbours and $\mathcal{Q}$ is an
optional set of operator-defined key points, with $s_j\in\{+1,-1\}$ determining attraction or repulsion. Aggregation and diffusion use a constant neighbour weighting, $f(d_j)=1/k$ for $k$ neighbours, producing motion toward or away from the local centroid. Lennard--Jones interaction instead uses

\begin{equation}
f_{LJ}(d) \;=\; \frac{\epsilon}{d}
\left[
\left(\frac{\delta}{d}\right)^4 -
\left(\frac{\delta}{d}\right)^2
\right],
\label{eq:lj}
\end{equation}

for which the attractive and repulsive contributions cancel at $d=\delta$.
Consequently, $\delta$ specifies the target inter-agent spacing and $\epsilon$ controls the strength of the interaction. Operator-defined points enter Eq.~\ref{eq:potential} through $\mathcal{Q}$, with their sign and gain determining attraction or repulsion. These inputs generate the Attraction point and Repulsion point behaviours and can also be superimposed on Pursuit to displace the circulating formation. The resulting motion command is saturated at the platform limit $v_{\max}$ before being passed to the platform adapter.

\paragraph{Cyclic pursuit}
Pursuit is generated by a distinct controller. Linear velocity $v$ remains positive and constant, while angular velocity switches according to whether another robot is detected ahead:

\begin{equation}
\omega =
\begin{cases}
\omega^{+}, & \text{if a neighbour is detected ahead},\\[2pt]
\omega^{-}, & \text{otherwise},
\end{cases}
\label{eq:pursuit}
\end{equation}

with $\omega^{+}>0>\omega^{-}$. The robots converge toward a circulating formation whose characteristic radius depends on $v$ and the angular-velocity parameters ~\cite{pursuit2018}. These parameters were adjusted to the available performance space so that circulation could be realised both on the approximately metre-wide Sushi tabletop and on stages several metres across.

\paragraph{Composition and synchronised transitions}
The remaining behaviours combine or sequence the preceding controllers. Fireworks alternates aggregation and diffusion using a counter $\tau$ that advances once per control step. A transition occurs when the counter reaches a prescribed threshold $T$, producing repeated contraction and expansion of the group. Thus, $T$ specifies the number of controller iterations assigned to each phase rather than a duration in physical time; the realised period consequently depends on the execution frequency of the platform. The diffusion gain controls the spatial extent reached during the expansion phase.

Longer sequences, including Fireworks Spiral and Fireworks Curve, schedule successive controller configurations. Fireworks + Pursuit instead combines simultaneous execution of the two controllers by separate subgroups. Transitions between states are synchronised through a Buzz barrier: each robot signals readiness, and the swarm advances only once the barrier condition is satisfied. This prevents individual robots from entering the next controller state independently and preserves each transition as a group-level event.

\paragraph{Deployment and live control}
During performances, each swarm was operated through a ground-station interface used to launch, stop, and switch behaviours and to modify the subset of parameters exposed for live interaction. These included external attraction or repulsion points and the platform velocity limit; controller-internal parameters were fixed during rehearsal unless otherwise specified. The interface therefore allowed the operator to select behaviours and alter predefined external inputs without modifying the underlying Buzz program during a performance.

% \paragraph{Parameters and platform dependence}
% \todonotes[inline]{ALI: add a table with actual parameters used for the 3 kernels defined above.}

% \todonotes[inline]{Describe the practical deployment procedure: simulation tests, parameter tuning, rehearsal runs, transfer to physical robots, safety checks, operator responsibilities, logging, and fallback/manual intervention procedures.}

\subsection{Axis 2: The interpretation step}\label{sec:axis2}

The perceptual axis treated the discrepancy between realised collective organisation and human interpretation as an iterative design problem rather than as a single performance metric. Behaviour, transitions, embodiment, scenography, sound, appearance, performers, and audience mediation were co-developed. Because these components changed together across the research-creation process, DESSAIM does not isolate the causal contribution of any one component. Instead, the method used repeated observation, notation, interpretation, and redesign to make perceptual mismatches actionable for artists and roboticists.

\subsubsection{Research-creation and iterative design protocol}

\begin{figure}[h!]
  \centering
  \includegraphics[width=0.85\columnwidth]{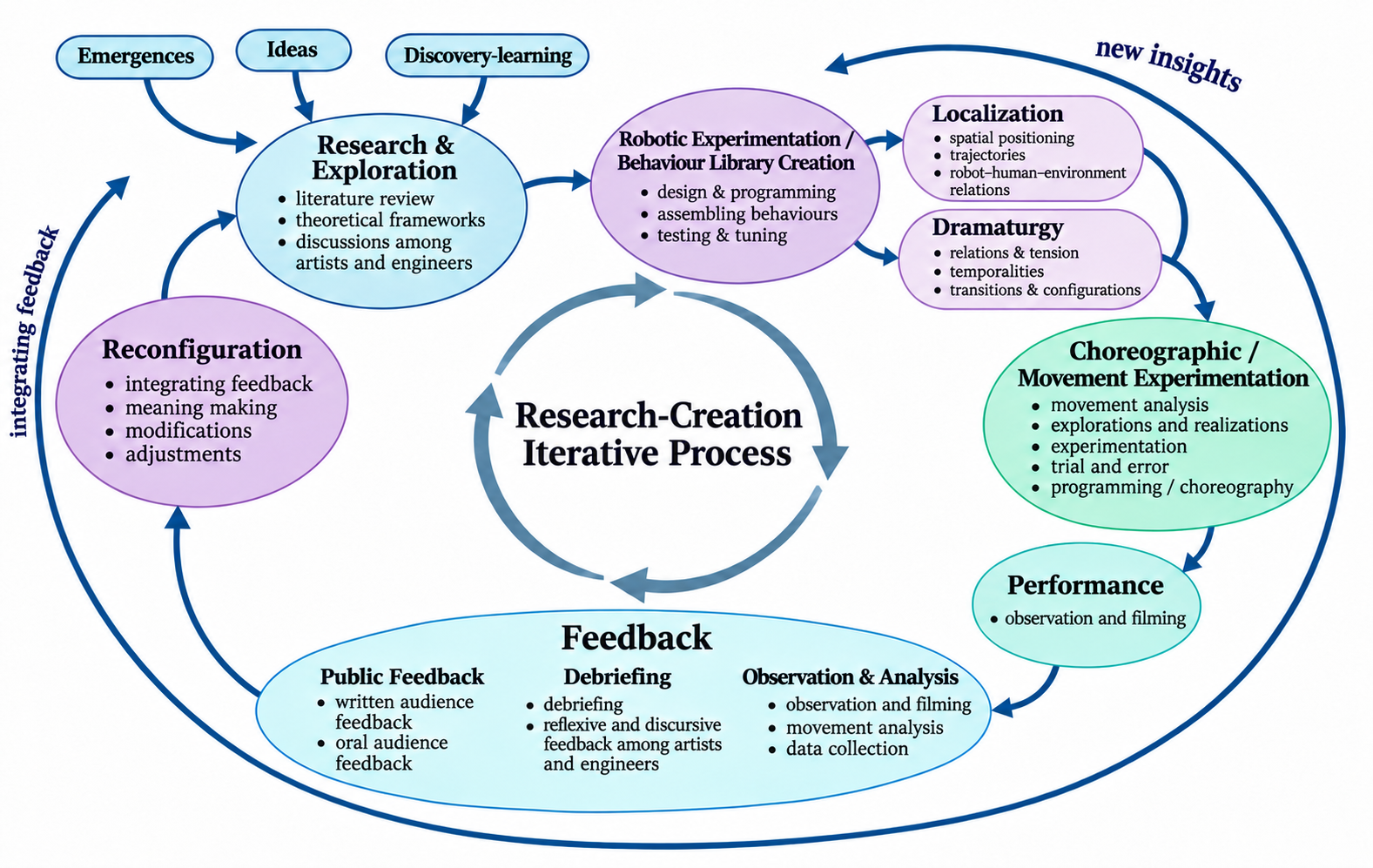}
  \caption{Iterative research-creation process used in DESSAIM. Robotic experimentation, choreographic and movement exploration, performance, observation, public feedback, and interdisciplinary debriefing were linked through repeated cycles of reconfiguration. Technical, artistic, and perceptual observations could be reinvested in subsequent iterations, allowing changes to robot behaviour, localisation, dramaturgy, staging, embodiment, and mediation.}
  \label{fig:research_creation_process}
\end{figure}

DESSAIM took place between 2022 and 2024. Preparatory work involving artists and engineers was conducted before and between four two-week laboratory residencies in large performance spaces in Montreal. The residencies combined robotic development, movement exploration, scenographic composition, rehearsal, and public presentation.

Ten public presentations were held during the project, including four presentations during the Casteliers Festival in March 2024. The public presentations were walk-through performance-installations rather than seated demonstrations: audiences moved within defined areas around the robotic systems and encountered the different swarms from changing viewpoints.

The research-creation process followed an iterative cycle (Fig.~\ref{fig:research_creation_process}). Emerging technical and artistic propositions were explored during each residency, followed by rehearsal or public presentation, observation, interdisciplinary debriefing, and reconfiguration. Informal public impressions, technical observations, and artist--engineer discussions could lead to changes in swarm behaviour, behavioural transitions, localisation, dramaturgy, arm posture, sound, materials, staging, audience viewpoint, or mediation before the next iteration. Public feedback therefore informed subsequent design decisions rather than serving only as an end-of-project evaluation.

\subsubsection{Movement Observation--Analysis as an iterative method}

Movement Observation–Analysis (MOA; OAM in French)~\cite{harbonnier2021} was used as a shared observation and notation framework between choreographic and engineering practices. MOA draws from Functional Analysis of the Dancing Body (AFCMD) and Laban–Bartenieff Movement Analysis (LBMA) and organises movement observations around Ground, Space, and Dynamics.

MOA proceeds through three interconnected stages. First, observers allow subjective impressions to emerge through perception and kinaesthetic engagement. Second, these impressions are examined through a structured description of observable movement parameters. Third, the observations are brought together in a construction of meaning that links perceived qualities to the movement organisation that may have produced them.

Within DESSAIM, this three-stage process was extended into an iterative design loop. Initial impressions arising during laboratories and public presentations were reformulated as observable movement descriptors; these descriptors were then connected to variables that could be modified in the swarm behaviour, robot posture, or staging; the revised configuration was subsequently rehearsed or presented again and observed anew.

For example, observations of the Sushis could distinguish rapid changes of direction, short stops, centrifugal and centripetal motion, variations in velocity, and changes in group density. These descriptors provided a vocabulary through which choreographers and roboticists could discuss possible interventions in velocity limits, attraction/repulsion parameters, inter-agent spacing, behavioural sequence, or timing. MOA was therefore used as a notation and convergence tool rather than as a deterministic mapping from an artistic quality to a numerical control parameter. Representative excerpts from the public performances and research-creation laboratories are provided in the performance videos online \footnote{\url{https://initrobots.ca/dessaim/}}.

Comments and observations gathered during laboratories and debriefings were used as design inputs rather than treated as part of the formally coded audience dataset; recurrent observations concerning changes in speed, rhythm, and collective organisation contributed to the later emphasis on transitions between collective states.

%\todonotes[inline]{Describe MOA/OAM as the framework used to translate between movement qualities and swarm parameters. Include the three-step process from the presentation: first subjective impressions, analysis of movement parameters, and construction of meaning. Mention CORA-SC only if it is necessary to formalize the mapping.}

%\todonotes[inline]{Describe how choreographic and puppetry intentions mapped to controllable features: spatial relation, density, tempo, expansion/contraction, attraction/repulsion, posture gesture for Doodies, point of view, scale, shadow, sound, costumes, and scenography.}

\subsubsection{Behaviour composition and transition design}
Movement sequences combined pre-programmed swarm behaviours with operator-triggered changes and constrained improvisational elements. The twelve behaviours listed in Table~\ref{tab:behaviors} provided a behavioural vocabulary from which sequences were assembled. Operators could trigger behaviour changes and modify attraction or repulsion points while the underlying decentralised Buzz program remained unchanged. Example behaviour timelines for two public performances are shown in Supplementary Fig.~B

Composition focused not only on steady-state behaviours but also on changes between collective regimes. Aggregation, diffusion, pursuit, fireworks, and attraction/repulsion were combined to create changes in density, direction, circulation, spacing, and organisation. Spatial relationships, orientation relative to gravity, speed, acceleration, expansion and contraction, and—on the Doodies—the relationship between arm posture and base motion were treated as compositional variables.

One Sushi sequence, for example, progressed through aggregation, diffusion, and pursuit on an oval Plexiglas surface beneath a cave-like scenographic structure. Other sequences alternated attraction and dispersion around a point or along trajectories governed by Lennard–Jones interactions. These examples are included here to specify how the swarm algorithms were composed into performance sequences; audience interpretations of these sequences are reported in Results.

%\todonotes[inline]{Explain how behaviours were composed into readable sequences. Include ARGoS composition, aggregation--diffusion--pursuit sequences, fireworks behaviours, operator-triggered attraction/repulsion points, and any teleoperation/recording/direct manipulation procedures already present in the draft.}

%\todonotes[inline]{Emphasize transitions as methodological objects: humans often read aim through changes between collective states rather than through a steady-state swarm pattern alone.}

\subsubsection{Embodiment and the extended expressive toolkit}

The swarm behaviours were presented within scenographic environments designed to modify how collective motion could be perceived. The expressive design space included robot morphology, coverings, recycled materials, lighting, projected shadows, amplified mechanical sound, spatial relationships with performers, and spectators’ viewpoints.

Recycled textiles, VHS tape, moulded packaging, cardboard and feathers were assembled into lightweight coverings and scenographic structures such as skirts, wings, screens and curtains. These materials were selected and modified during the iterative process to change apparent weight, scale, continuity of shape, and the visibility of small movements.

Light and shadow were used to extend or transform the apparent scale of the swarms. For the CrazyCognies, airflow also animated lightweight scenographic materials such as VHS strips, allowing the physical effects of the robots’ motion to become visible beyond the vehicle bodies themselves. Contact microphones and electromagnetic microphones were used to amplify mechanical and motor sounds when these were incorporated into the performance composition.

Embodiment also shaped the available movement vocabulary. The Sushi movement vocabulary was composed primarily through planar position, density, speed, and interaction with tabletop scenography. The CrazyCognies added vertical motion and airflow. The Doodies combined planar base trajectories with multi-degree-of-freedom arm motion. In Doodies Divas, for example, fabric skirts covered the industrial bases while the articulated arms moved in relation to the swarm behaviour. Arm posture and motion could precede, accompany, or contrast with base displacement and were therefore treated as additional compositional variables. Audience interpretations of these embodiments are reported in Results rather than here.

%\todonotes[inline]{Describe set design as part of the perceptual method. Include recycled materials, costumes, shadows, screens/curtains, wings, VHS strips, nests/grottes/rock-like forms, lighting, sound capture, contact microphones/electromagnetic pickups, and mediation strategy where applicable.}

%\todonotes[inline]{Explain why these elements matter for robotics: they are not only aesthetic additions; they are perceptual channels that help or hinder the reading of collective aim.}

\subsubsection{Audience questionnaire and qualitative data collection}\label{sec:qual}

Following public presentations, electronic questionnaires were administered using tablets available at the venue or QR codes accessed on participants’ personal devices. The study was approved by the ÉTS ethics committee under protocol \emph{H20220301}, and participants provided informed consent before completing the questionnaire. Follow-up emails were used to encourage participation. A participant could contribute to only one performance evaluation.

For each of the three robotic systems—Sushis, CrazyCognies, and Doodies—participants rated eleven statements addressing sympathy, emotional response, indifference, hostility, meaningfulness of motion, destabilization, appreciation of form, comfort, impressiveness, influence of spatial staging, and influence of appearance or costume. Responses were recorded on a seven-level agreement scale encoded from 0 (strongly disagree) to 6 (strongly agree). Two open-ended questions asked what the robots evoked and what they appeared to communicate.

The analysis included 54 complete responses: 13 for the first performance, 12 for the second, and 29 for the third. Each participant evaluated all three robotic systems, enabling paired comparisons between embodiments, whereas comparisons across performances involved independent participant cohorts. Open-ended responses were retained in their original language and organised by robotic system, performance, and prompt.

The questionnaire was not designed as a direct test of whether participants correctly identified a predefined swarm goal. In the present paper, the meaningfulness item and the open-ended communication responses are treated as indicators of meaning attribution, while sympathy, hostility, comfort, and destabilization are treated as affective responses. Appreciation of form, impressiveness, and the staging/appearance items characterise other perceptual dimensions of the expressive design space. These constructs are related but are not treated as interchangeable measures of the perceptual gap.

%\todonotes[inline]{Provide the audience-study protocol here, not in Results. Include questionnaire wording, response scale, languages, administration method, inclusion/exclusion criteria if any, number of responses per restitution/performance, whether participants could return across sessions, and missing-data handling.}

%\todonotes[inline]{Describe written impressions, oral discussions, group discussions, QR-code questionnaire, and how public feedback was reinvested into later laboratories.}

\subsubsection{Quantitative and qualitative analysis}\label{sec:quant}

Likert ratings were treated as ordinal variables and summarised using medians and
interquartile ranges. For each of the 11 questionnaire items, a cumulative-link
mixed model with a logit link was fitted with performance (three levels), robotic
embodiment (CrazyCognies, Sushis, and Doodies), and their interaction as fixed
effects, and participant as a random intercept. Performance was a between-participant
factor because each participant attended only one performance, whereas embodiment
was a within-participant factor because each participant evaluated all three robotic
systems.

Omnibus effects of performance, embodiment, and their interaction were evaluated
using Wald $\chi^2$ tests. To control for multiple testing across questionnaire
dimensions, Holm correction was applied separately to three families:
the 11 performance effects, the 11 embodiment effects, and the 11
performance-by-embodiment interactions. Pairwise model contrasts were used to
localise relevant effects and were Holm-corrected within each questionnaire item
and contrast family. Effect estimates are reported as cumulative odds ratios (OR)
with 95\% confidence intervals. Complete descriptive statistics and model results
for all 11 questionnaire items, including non-significant outcomes and post-hoc
contrasts, are reported in Supplementary Table~A. Given the unequal performance
cohort sizes and uncontrolled differences between public performance contexts,
inferential results were interpreted as exploratory.

Open-ended responses were analysed using exploratory thematic coding. Two
researchers independently examined and coded the responses, compared their
analyses, and resolved differences through discussion. Coding was performed separately for the prompts asking what the robots evoked and what they appeared to communicate, and was organised by robotic system and performance. Recurrent themes included biological or organic
associations, collective and social organisation, technological imagery,
anthropomorphic or zoomorphic interpretations, agency, affiliation, and threat.
Responses could receive multiple codes when several ideas were present. The
resulting themes were compared across robotic systems and performances and
subsequently interpreted alongside the quantitative results.

%%%%%%%%%%%%%%%%%%%%%%%%%%%%%%%%%%%%%%%%%%%%%%%%%%%%%%%%%%%%%%%%%%%%%%
% \section{Conclusion}
%%%%%%%%%%%%%%%%%%%%%%%%%%%%%%%%%%%%%%%%%%%%%%%%%%%%%%%%%%%%%%%%%%%%%%

%\todonotes[inline]{Conclude in one or two concise paragraphs. Return to the central thesis: swarm robotics inherits the reality gap from simulation to deployment, but public performance revealed a second perceptual gap between what the collective does and what humans read as aim. DESSAIM suggests that expressive swarms require both robust deployment infrastructure and a full expressive toolkit.}

%\todonotes[inline]{Include future work here rather than as a Discussion subsection: controlled comparison of individual versus collective cues; systematic study of transitions; mapping MOA parameters to swarm-control parameters; cross-platform replication by other groups; larger swarms; longitudinal audience perception; and formal links between technical robustness and perceived aim.}

%%%%%%%%%%%%%%%%%%%%%%%%%%%%%%%%%%%%%%%%%%%%%%%%%%%%%%%%%%%%%%%%%%%%%%
% Backmatter blocks (npj expects these as dedicated statements)
%%%%%%%%%%%%%%%%%%%%%%%%%%%%%%%%%%%%%%%%%%%%%%%%%%%%%%%%%%%%%%%%%%%%%%
\backmatter

\bmhead{Data availability}
Sample robot logs from public performances, localisation logs, communication logs and analysis scripts are available in the repository identified below. The study's ethics approval does not permit public release of the audience-response data.

\bmhead{Code availability}
PyBuzz modules, Buzz behaviour scripts, ARGoS simulations, platform adapters, configuration files, and setup documentation are available at \url{https://git.initrobots.ca/swarmui/buzz_unified}.

\bmhead{Acknowledgments}
The authors would like to acknowledge the creative and empathetic contribution of other talented artists and engineers of DESSAIM: Marcelle Hudon, Magali Babin, Rachid Aissaoui, Danielle Lecourtois, Corentin Boucher, Alexandra Mercader, Hao-Fang Cheng, Clement Gassmann-Prince, Guillaume Ricard and Radhouen Khmiri. This project would not have been possible without the financial support of the FRQSC Network cluster Hexagram and the FRQ Grant \emph{P}RISME|Art.

\bmhead{Author contributions}
D.S. developed the localization stack for the Doodies. D.S. and Y.B. developed the localisation stack for the CrazyCognies. G.B. and V.S.V. configured the communication network for the swarm deployments.
H.D., A.R. and J.M. created the performances, including their choreography and scenography, with H.D. leading the artistic team.
A.I., A.B.M.A., R.G.B., V.S.V., Y.B. and J.M. were in charge of the robotic systems over the all rehearsal and public performances. R.G.B. and V.S.V. ran the motion-capture sessions, and R.G.B., A.B.M.A. and V.S.V. ran the Sushi simulations.
A.I. analysed the localisation and trajectory data and generated their visualisation. D.S., H.D. ran the audience study, analysed the ratings and coded the open-ended responses.
R.G.B., V.S.V, A.B.M.A. and A.I. built PyBuzz on top of the Buzz and implemented the platform adapters.
A.I., A.B.M.A, D.S. and G.B. conceived the perceptual-gap framework.
A.I., A.B.M.A., R.G.B., H.D. and D.S. wrote the manuscript. D.S. led the DESSAIM project and acquired its funding; D.S. and G.B. supervised the work and revised the manuscript. All authors reviewed and approved the final manuscript.

\bmhead{Competing Interests}
All authors declare no financial or non-financial competing interests. 

%%%%%%%%%%%%%%%%%%%%%%%%%%%%%%%%%%%%%%%%%%%%%%%%%%%%%%%%%%%%%%%%%%%%%%
%% References
%%%%%%%%%%%%%%%%%%%%%%%%%%%%%%%%%%%%%%%%%%%%%%%%%%%%%%%%%%%%%%%%%%%%%%
\bibliographystyle{sn-nature}
\bibliography{references,expmot,bibliography}

\end{document}